\documentclass[journal]{IEEEtran}
\usepackage{cite}

\ifCLASSINFOpdf
  \usepackage[pdftex]{graphicx}
  \graphicspath{{../pdf/}{../jpeg/}}
  \DeclareGraphicsExtensions{.pdf,.jpeg,.png}
\else
  \usepackage[dvips]{graphicx}
  \graphicspath{{../eps/}}
  \DeclareGraphicsExtensions{.eps}
\fi
\usepackage{amsmath}

\usepackage{array}

\ifCLASSOPTIONcompsoc
 \usepackage[caption=false,font=normalsize,labelfont=sf,textfont=sf]{subfig}
\else
 \usepackage[caption=false,font=footnotesize]{subfig}
\fi

\usepackage{dblfloatfix}
\usepackage{url}
\usepackage{amssymb}
\usepackage{amsthm}
\usepackage{bm}
\usepackage{mathtools}
\usepackage{dsfont}

\usepackage{url}
\usepackage{color,soul}

\newcommand{\R}[0]{\mathds{R}} % real numbers
\newcommand{\C}[0]{\mathds{C}} % complex numbers
\providecommand{\mb}[1]{\mathbf{#1}}
\providecommand{\mbb}[1]{\boldsymbol{#1}}
\providecommand{\mbx}{\mb{x}}
\providecommand{\mby}{\mb{y}}

\providecommand{\pfpx}[2]{\frac{\partial{#1}}{\partial{#2}}}
\providecommand{\dft}{\mb{F}}
\providecommand{\idft}{\mb{F}^{-1}}
\providecommand{\kx}{\mb{\hat{x}}}

\begin{document}

\IEEEpubid{\begin{minipage}{\textwidth}\ \\[12pt]
    Copyright (c) 2017 IEEE. Personal use of this material is permitted. 
    However, permission to use this material for any other purposes must be
    obtained from the IEEE by sending a request to pubs-permissions@ieee.org
\end{minipage}}

\title{A Deep Cascade of Convolutional Neural Networks for Dynamic MR Image Reconstruction}

\author{Jo~Schlemper*, Jose~Caballero, Joseph~V.~Hajnal, Anthony~Price~and~Daniel~Rueckert,~\IEEEmembership{Fellow,~IEEE}%
\thanks{*J. Schlemper is with the Department of Computing, Imperial College London, SW7 2AZ London, U.K. (e-mail: jo.schlemper11@imperial.ac.uk).}%
\thanks{A. N. Price and J. V. Hajnal are with the Division of Imaging Sciences and Biomedical Engineering Department, King's College London, St. Thomas' Hospital, SE1 7EH London, U.K. (email: anthony.price@kcl.ac.uk; jo.hajnal@kcl.ac.uk).}%
\thanks{J. Caballero and D. Rueckert is with the Department of Computing, Imperial College London, SW7 2AZ London, U.K. (e-mail: jose.caballero06@imperial.ac.uk; d.rueckert@imperial.ac.uk).}%
}

\markboth{Accepted for IEEE Transactions on Medical Imaging}%
{Schlemper \MakeLowercase{\textit{et al.}}: A Deep Cascade of Convolutional Neural Networks for Dynamic MR Image Reconstruction}


% \IEEEpubid{Copyright (c) 2017 IEEE. Personal use of this material is permitted. 
%   However, permission to use this material \\ for any other purposes must be
%   obtained from the IEEE by sending a request to pubs-permissions@ieee.org}



\maketitle

\begin{abstract}

Inspired by recent advances in deep learning, we propose a framework for reconstructing dynamic sequences of 2D cardiac magnetic resonance (MR) images from undersampled data using a deep cascade of convolutional neural networks (CNNs) to accelerate the data acquisition process. In particular, we address the case where data is acquired using aggressive Cartesian undersampling. Firstly, we show that when each 2D image frame is reconstructed independently, the proposed method outperforms state-of-the-art 2D compressed sensing approaches such as dictionary learning-based MR image reconstruction, in terms of reconstruction error and reconstruction speed. Secondly, when reconstructing the frames of the sequences jointly, we demonstrate that CNNs can learn spatio-temporal correlations efficiently by combining convolution and data sharing approaches. We show that the proposed method consistently outperforms state-of-the-art methods and is capable of preserving anatomical structure more faithfully up to 11-fold undersampling. Moreover, reconstruction is very fast: each complete dynamic sequence can be reconstructed in less than 10s and, for the 2D case, each image frame can be reconstructed in 23ms, enabling real-time applications.

\end{abstract}


% \fi
\IEEEpeerreviewmaketitle

\section{Introduction}

\IEEEPARstart{I}n many clinical scenarios, medical imaging is an indispensable diagnostic and research tool.
One such important modality is Magnetic Resonance Imaging (MRI), which is
non-invasive and offers excellent resolution with various contrast mechanisms to
reveal different properties of the underlying anatomy. However, MRI is associated
with an inherently slow acquisition process. This is because data samples of an MR image are acquired sequentially in \emph{$k$-space} and the speed at which $k$-space can be traversed is limited by physiological and hardware constraints \cite{Lustig2008}. A long data acquisition procedure imposes significant demands on patients, making this imaging modality expensive and less accessible. One possible
approach to accelerate the acquisition process is to undersample $k$-space, which in theory
provides an acceleration rate proportional to a reduction factor of a number of $k$-space traversals required. However,
undersampling in $k$-space violates the Nyquist-Shannon theorem and generates
aliasing artefacts when the image is reconstructed. The main challenge in this case is to find an algorithm that can recover an uncorrupted image taking into account the undersampling regime combined with a-priori knowledge of appropriate properties of the image to be reconstructed.

Using Compressed Sensing (CS), images can be reconstructed from sub-Nyquist
sampling, assuming the following: firstly, the images must be \emph{compressible}, i.e.
they have a sparse representation in some transform domain. Secondly, one must ensure \emph{incoherence} between the sampling and sparsity domains to guarantee that the reconstruction problem has a unique solution and that this solution is attainable. In practice, this can be achieved with random subsampling of $k$-space, which produces aliasing patterns in the image domain that can be regarded as correlated noise. Under such assumptions, images can then be reconstructed through nonlinear optimisation or iterative algorithms. The class of methods which apply CS to the MR reconstruction problem is termed CS-MRI \cite{Lustig2008}. In general, these methods use a fixed sparsifying transforms, e.g. wavelet transformations. A natural extension of these approaches has been to enable more flexible representations with \emph{adaptive} sparse modelling, where one attempts to learn the optimal sparse representation from the data directly. This can be done by exploiting, for example, dictionary learning (DL) \cite{Ravishankar2011}. 

To achieve more aggressive undersampling, several strategies can be considered. One way is to further exploit the inherent redundancy of the MR data. For example, in dynamic imaging, one can make use of spatio-temporal redundancies \cite{Caballero2014}, \cite{Jung2007}, \cite{Quan2016}, or when imaging a full 3D volume, one can exploit redundancy from adjacent slices \cite{Hirabayashi2015}. An alternative approach is to exploit sources of explicit redundancy of the data to turn the initially underdetermined problem arising from undersampling into a determined or overdetermined problem that is easily solved. This is the fundamental assumption underlying parallel imaging \cite{Uecker2014}. Similarly, one can make use of multi-contrast information 
\cite{Huang2014a} or the redundancy generated by multiple filter responses of the image \cite{XiPeng2015}. These explicit redundancies can also be used to complement the sparse modelling of inherent redundancies \cite{Jin2015}, \cite{Liang2009a}.

 \begin{figure*}[t]
  \centering
  \includegraphics[width=1\textwidth]{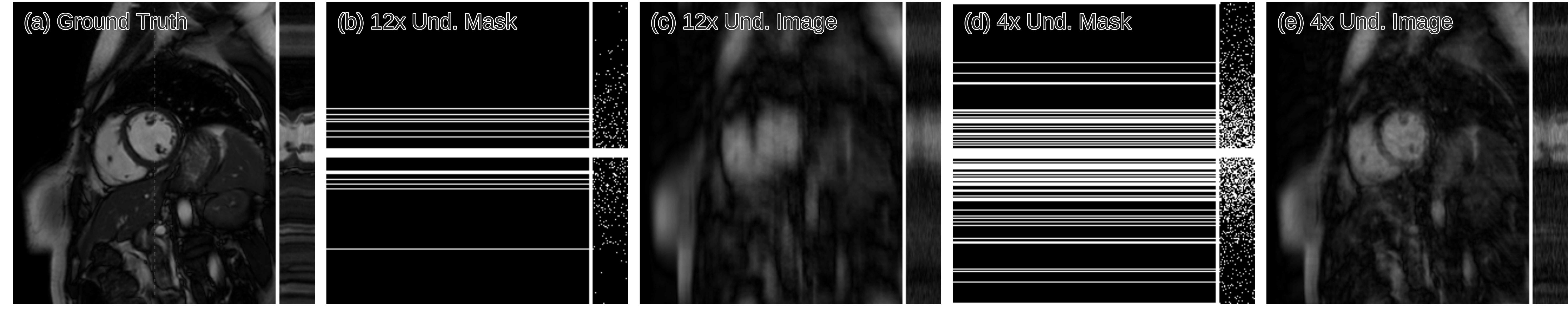}
  \caption{An example of the image acquisition with Cartesian undersampling for a sequence of cardiac cine images. (a) A ground truth sequence that is fully-sampled in $k$-space, shown along $x$-$y$ and $y$-$t$ for the image frame and the temporal profile respectively. (b) A Cartesian undersampling mask that only acquires 1/12 of samples in $k$-space, where white indicates the sampled lines. Each image frame is undersampled with the mask shown along $k_x$-$k_y$. The undersampling pattern along the temporal dimension is shown in $k_y$-$t$. (c) The zero-filled reconstruction of the image acquired using the 12-fold undersampling mask. (d, e) 4-fold Cartesian undersampling mask and the resulting zero-filled image. Note that the aliasing artefact becomes more prominent as the undersampling factor is increased.}
\label{fig:undersampling_example} 
\end{figure*}

\IEEEpubidadjcol
Recently, deep learning has been successful at tackling many computer vision problems. Deep neural network architectures, in particular convolutional
neural networks (CNNs), are becoming the state-of-the-art technique for various
imaging problems including image classification \cite{He2015}, object localisation \cite{Ren2015} and image
segmentation \cite{Ronneberger2015}.  Deep architectures are capable of extracting features from data to build increasingly abstract representations, replacing the traditional approach of carefully hand-crafting features and algorithms. For example, it has already been demonstrated that CNNs outperform sparsity-based methods in super-resolution \cite{Dong2016} in terms of both reconstruction quality and speed \cite{Shi_2016_CVPR}. One of the contributions of our work is to explore the application of CNNs in undersampled MR reconstruction and investigate whether they can exploit data redundancy through learned representations. In fact, CNNs have already been applied to compressed sensing from random Gaussian measurements
\cite{Kulkarni2016a}. Despite the popularity of CNNs, there has only been preliminary research on CNN-based MR image reconstruction \cite{NIPS2016_6406}, \cite{wang2016b},  hence the applicability of CNNs to this problem for various imaging protocols has yet to be fully explored. 

In this work we consider reconstructing dynamic sequences of 2D cardiac MR images with Cartesian undersampling, as well as reconstructing each frame independently, using CNNs. We view the reconstruction problem as a de-aliasing problem in the image domain. Reconstruction of undersampled MR images is challenging because the images
typically have low signal-to-noise ratio, yet often high-quality
reconstructions are needed for clinical applications. To resolve this issue, we propose a deep network architecture which forms a cascade of
CNNs.\footnote{Code available at \url{https://github.com/js3611/Deep-MRI-Reconstruction}} Our cascade network closely resembles the iterative reconstruction of DL-based methods, however, our approach allows end-to-end optimization of the reconstruction algorithm. 
For 2D reconstruction, the proposed method is compared to \emph{Dictionary Learning MRI (DLMRI)} \cite{Ravishankar2011} and for dynamic reconstruction, the method is compared to Dictionary Learning with
Temporal Gradient (\emph{DLTG}) \cite{Caballero2014}, kt Sparse and Low-Rank (kt-SLR) \cite{lingala2011} and \emph{Low-Rank Plus Sparse Matrix Decomposition (L+S)} \cite{otazo2014}, which are the state-of-the-art compressed sensing and low-rank approaches. We show that the proposed method outperforms them in terms of reconstruction error and perceptual quality, especially for aggressive undersampling rates. Moreover, owing to GPU-accelerated libraries, images can be reconstructed efficiently using the approach. In particular, for 2D reconstruction, each image can be reconstructed in about 23ms, which is fast enough to enable real-time applications. For the dynamic case, sequences can be reconstructed within 10s, which is reasonably fast for off-line reconstruction methods.

\section{Problem Formulation}

Let $\mbx \in \C^N$ represent a sequence of 2D complex-valued MR images stacked as a column vector, where $N= N_x N_y N_t$. Our problem is to reconstruct $\mbx$ from $\mby \in \C^M$ ($M \ll N$), undersampled measurements in $k$-space, such
that:

\begin{equation}
  \mby = \mb{F}_u \mbx + \mb{e}
  \label{eq:cs_basic}
\end{equation}

Here $ \mb{F}_u \in \C^{M\times N}$ is an undersampled Fourier encoding matrix and $\mb{e} \in \C^M$ is acquisition noise modelled as additive white Gaussian (AWG) noise. In the case of Cartesian acquisition, we have $\mb{F}_u = \mb{M}\mb{F}$, where $\mb{F}  \in \C^{N\times N}$ applies two-dimensional Discrete Fourier Transform (DFT) to each frame in the sequence and $\mb{M}  \in \C^{M\times N}$ is an undersampling mask selecting lines in $k$-space to be sampled for each frame. The corresponding subset of indices sampled in $k$-space is indicated by $\Omega$. For the fully-sampled case, $M=N$, the sequence is reconstructed by applying the 2D inverse DFT (IDFT) to each frame. However, Eq. (\ref{eq:cs_basic}) is underdetermined even in the absence of noise, and hence the inversion is ill-posed; in particular, applying IDFT, which in this case is also called \emph{zero-filled} reconstruction, results in a sequence of aliased images $\mbx_u = \mb{F}^{H}_u \mby$ due to sub-Nyquist sampling. Note that $\mb{F}^{H}_u$ is the Hermitian of the encoding matrix, which first maps $\mby \in \C^M$ to the $k$-$t$ coordinate and then applies the 2D IDFT frame-wise.  Examples of the aliased images are shown in Fig. \ref{fig:undersampling_example}. Therefore, in order to reconstruct $\mbx$, one must exploit a-priori knowledge of its properties, which can be done by formulating an unconstrained optimisation problem:

\begin{equation}
  \label{eq:sparse_coding}
\begin{aligned}
& \underset{\mbx}{\text{min.}}
& & \mathcal{R}(\mbx) + \lambda \| \mby - \mb{F}_u \mbx \|^2_2
\end{aligned}
\end{equation}

$\mathcal{R}$ expresses regularisation terms on $\mbx$ and $\lambda \in \R$ allows the adjustment of data fidelity based on the noise level of the acquired
measurements $\mby$. For CS-based methods, the regularisation terms $\mathcal{R}$ typically involve $\ell_0$ or
$\ell_1$ norms in the sparsifying domain of $\mbx$. Our formulation is inspired by DL-based reconstruction approaches \cite{Ravishankar2011}, in which the problem is formulated as:

\begin{equation}
  \label{eq:dl}
\begin{aligned}
& \underset{\mbx, \mb{D}, \{ \mbb{\gamma}_i\} }{\text{min.}}
& & \sum_i \left(\|\mb{R}_i\mb{x} - \mb{D} \mbb{\gamma}_i \|_2^2 + \nu \|\mbb{\gamma}_i \|_0 \right) + \lambda \| \mby - \mb{F}_u \mbx \|^2_2
\end{aligned}
\end{equation}

Here $\mb{R}_i$ is an operator which extracts a spatio-temporal image patch at $i$, $\mbb{\gamma}_i$ is the corresponding sparse code with respect to a dictionary $\mb{D}$. In this approach, the regularisation terms force $\mbx$ to be approximated by the reconstructions from the sparse code of patches. By taking the same approach, for our CNN formulation, we force $\mb{x}$ to be well-approximated by the CNN reconstruction:

\begin{equation}
  \begin{aligned}
    \label{eq:cnn_rec}
    & \underset{\mbx}{\text{min.}}
    & & \| \mbx - f_{\text{cnn}}(\mbx_u | \boldsymbol{\theta}) \|^2_2 + \lambda \| \mb{F}_u \mbx - \mby \|_2^2 \\ 
  \end{aligned}
\end{equation}

Here $f_{\text{cnn}}$ is the forward mapping of the CNN parameterised by $\boldsymbol{\theta}$, possibly containing millions of adjustable network weights, which takes in the zero-filled reconstruction $\mbx_u$ and directly produces a reconstruction as an output. Since $\mb{x}_u$ is heavily affected by aliasing from sub-Nyquist sampling, the CNN reconstruction can therefore be seen as solving a de-aliasing problem in the image domain. The approach of Eq. (\ref{eq:cnn_rec}), however, is limited in the sense that the CNN reconstruction and the data fidelity are two independent terms. In particular, since the CNN operates in the image domain, it is trained to reconstruct the sequence without a-priori information of the acquired data in $k$-space. However,
if we already know some of the $k$-space values, then the CNN should be discouraged from modifying them, up to the level of acquisition noise. Therefore, by incorporating the data fidelity in the learning stage, the CNN should be able to achieve better reconstruction. This means that the output of the CNN is now conditioned on $\Omega$ and $\lambda$. Then, our final reconstruction is given simply by the output, $\mbx_\text{cnn} = f_{\text{cnn}}(\mbx_u | \boldsymbol{\theta}, \lambda, \Omega)$. Given training data
$\mathcal{D}$ of input-target pairs $(\mbx_u, \mbx_\text{gnd})$ where $\mbx_\text{gnd}$ is a fully-sampled ground-truth data, we can train the CNN to produce an output that attempts to accurately reconstruct the data by minimising an objective function:

\begin{equation}
\mathcal{L}(\boldsymbol{\theta}) = \sum_{(\mbx_u, \mbx_\text{gnd}) \in \mathcal{D}} \ell \left( \mbx_\text{gnd},
\mbx_\text{cnn} \right)\label{eq:cnn_objective}
\end{equation}

where $\ell$ is a loss function. In this work, we consider an element-wise squared loss, which is given by $\ell \left(\mbx_\text{gnd}, \mbx_\text{cnn} \right) = \| \mbx_\text{gnd} - \mbx_\text{cnn} \|_2^2$.

\section{Data Consistency Layer}

% Denote a Fourier transform of data $\mbx$ as $\hat{\mbx} = \mb{F}\mbx$, where $\mb{F}$ is the Fourier encoding matrix. In order to incorporate the data fidelity in the network architecture, we first note the following: for a fixed $\boldsymbol{\theta}$, Eq. (\ref{eq:cnn_rec}) has a closed-form solution $\hat{\mbx}_{\text{rec}}$ in $k$-space, given as in \cite{Ravishankar2011}:

% \begin{equation}
% \label{eq:dc_step_in_k}
% \hat{\mbx}_{\text{rec}}(k) =
%     \begin{cases}
%       \hat{\mbx}_{\text{cnn}}(k) & \text{if } k \not \in \Omega\\
%       \frac{\hat{\mbx}_{\text{cnn}}(k) + \lambda \hat{\mbx}_u(k)}{1+\lambda} & \text{if } k \in \Omega
%     \end{cases}
% \end{equation}

% where $\hat{\mbx}_{\text{cnn}} = \mb{F} f_{\text{cnn}}(\mbx_u | \boldsymbol{\theta})$. The final reconstruction is obtained by applying the inverse of the encoding matrix $\mbx_{\text{rec}} = \mb{F}^{-1}
% \hat{\mbx}_{\text{rec}}$. In the limit  $\lambda \to \infty$ we simply replace the $k$th predicted coefficient by the original coefficient if it
% has been sampled. For this reason, this operation is called a \emph{data consistency step} in $k$-space (DC). {\color{blue} In the case of the presence of non-neglegible noise during the acquisition, $\lambda$ must be adjusted accordingly (typically of order $10^3$ to $10^4$)}.

Denote the Fourier encoding of the image reconstructed by CNN as $\mb{s}_{\text{cnn}} = \mb{F}  \mbx_{\text{cnn}} = \mb{F} f_{\text{cnn}}(\mbx_u | \boldsymbol{\theta})$. $\mb{s}_{\text{cnn}}(j)$ represents an entry at index $j$ in $k$-space. The undersampled data $\mby \in \C^M$ can be mapped onto the vectorised representation of $k$-$t$ coordinate ($\C^N$) by $\mb{s}_0 = \mb{F} \mb{F}^H_u \mby$, which fills the non-acquired indices in $k$-space with zeros. In order to incorporate the data fidelity in the network architecture, we first note the following: for fixed network parameters $\boldsymbol{\theta}$, Eq. (\ref{eq:cnn_rec}) has a closed-form solution $\mb{s}_{\text{rec}}$ in $k$-space, given as in \cite{Ravishankar2011} element-wise:

\begin{equation}
\label{eq:dc_step_in_k}
\mb{s}_{\text{rec}}(j) =
    \begin{cases}
      \mb{s}_{\text{cnn}}(j) & \text{if } j \not \in \Omega\\
      \frac{\mb{s}_{\text{cnn}}(j) + \lambda \mb{s}_0(j)}{1+\lambda} & \text{if } j \in \Omega
    \end{cases}
\end{equation}

The final reconstruction in the image domain is then obtained by applying the inverse Fourier encoding $\mbx_{\text{rec}} = \mb{F}^{H}
\mb{s}_{\text{rec}}$. The solution yields a simple interpretation: if the $k$-space coefficient $\mb{s}_{\text{rec}}(j)$ is initially unknown (i.e. $j \not \in \Omega$), then we use the predicted value from the CNN. For the entries that have already been sampled ($j \in \Omega$), we take a linear combination between the CNN prediction and the original measurement, weighted by the level of noise present in $\mb{s}_0$. In the limit  $\lambda \to \infty$ we simply replace the $j$-th predicted coefficient in $\Omega$ by the original coefficient. For this reason, this operation is called a \emph{data consistency step} in $k$-space (DC). In the case of where there is non-neglegible noise present in the acquisition, $\lambda = q/\sigma$ must be adjusted accordingly, where $q$ is a hyper-parameter and $\sigma^2$ is the power of AWG noise in $k$-space (i.e. $\Re({\mb{e}_i}),\Im({\mb{e}_i}) \sim N(0, \sigma / \sqrt{2})$). In \cite{Caballero2014}, it is empirically shown that $p \in [5 \times 10^{-5}, 5\times 10^{-6}]$ for $\sigma^2 \in [4\times 10^{-8}, 10^{-9}]$ works sufficiently well.

Since the DC step has a simple expression, we can in fact treat it as a layer operation of the network, which we denote as a \emph{DC layer}. When defining a layer of a network, the rules for forward and backward passes must be specified in order for the network to be end-to-end trainable. This is because CNN training can effectively be performed through stochastic gradient descent, where one updates the network parameters $\boldsymbol{\theta}$ to minimise the objective function $\mathcal{L}$ by descending along the direction given by the derivative $\partial \mathcal{L} / \partial \boldsymbol{\theta}^T$. Therefore, it is necessary to define the gradients of each network layer relative to the network's output. In practice, one uses an efficient algorithm called \emph{backpropagation} \cite{backprop}, where the final gradient is given by the product of all the Jacobians of the layers contributing to the output. Hence, in general, it suffices to specify a layer operation $f_L$ for the forward pass and derive the Jacobian of the layer with respect to the layer input $\partial f_{L}
 / \partial \boldsymbol{\mbx}^T$ for the backward pass. 

\paragraph{Forward pass} The data consistency in $k$-space can be simply decomposed into three operations: Fourier transform $\mb{F}$, data consistency $f_{\text{dc}}$ and inverse Fourier transform $\mb{F}^H$. The data consistency $f_{dc}$ performs the element-wise operation defined in Eq. (\ref{eq:dc_step_in_k}), which, assuming $\mb{s}_0(j) = 0$ $\forall j \not \in \Omega$, can be written in matrix form as:

\begin{equation}
f_{dc}(\mb{s}, \mb{s}_0; \lambda) = \boldsymbol{\Lambda} \mb{s} + \frac{\lambda}{1 + \lambda} \mb{s}_0 \label{eq:fill_mat}
\end{equation}

Here $\bm{\Lambda}$ is a diagonal matrix of the form:

\begin{equation}
\bm{\Lambda}_{kk} =
\begin{cases}
  1 & \text{if } j \not \in \Omega \\
  \frac{1}{1+\lambda} & \text{if } j \in \Omega
\end{cases}\label{eq:fill_matrix}
\end{equation}

Combining the three operations defined above, we can obtain the forward pass of the layer performing data consistency in $k$-space: 

\begin{equation}
f_{L}(\mbx, \mby; \lambda) = \mb{F}^{H}\mb{\Lambda}\mb{F}\mbx + \frac{\lambda}{1 + \lambda} \mb{F}_u^{H}\mby\label{eq:dc_fnc}
\end{equation}

\paragraph{Backward pass} In general, one requires \emph{Wirtinger calculus} to derive a gradient in complex domain \cite{FaijulAmin2012}. However, in our case, the derivation greatly simplifies due to the linearity of the DFT matrix and the data consistency operation. The Jacobian of the DC layer with respect to the layer input $\mbx$ is therefore given by:
\begin{equation}
\pfpx{{f_{L}}}{\mbx^T} = \mb{F}^{H}\mb{\Lambda}\mb{F}
\end{equation}
Note that unlike many other applications where CNNs process real-valued data, MR images are complex-valued and the network needs to account for this. One possibility would be to design the network to perform complex-valued operations. A simpler approach, however, is to accommodate the complex nature of the data with real-valued operations in a dimensional space twice as large (i.e. we replace $\C^N$ by $\R^{2N}$).  In the latter case, the derivations above still hold due to the fundamental
assumption in Wirtinger calculus. 

The DC layer has one hyperparameter $\lambda \in \R$. This value can be fixed or made trainable. In the latter case, the derivative $\pfpx{ f_{\text{dc}} }{ \lambda }$ (a column vector here) is given by: 

\begin{equation}
\left[ \pfpx{f_{\text{dc}}(\mb{s}, \mb{s}_0; \lambda)}{\lambda} \right]_{j} =
\begin{cases}
  0 & \text{if } j \not \in \Omega \\
  \frac{\mb{s}_0(j)- \mb{s}_{\text{cnn}}(j)}{(1+\lambda)^2} & \text{if } j \in \Omega
\end{cases}
\label{eq:lambda_grad}
\end{equation}

and the update is $\Delta \lambda = \mb{J}_e \pfpx{ f_{\text{dc}} }{ \lambda }$ where $\mb{J}_e$ is the error backpropagated via the Jacobians of the layers proceeding $f_{\text{dc}}$.

\section{Cascading Network}

For CS-based methods, in particular for DL-based methods, the optimisation problem such as in Eq. (\ref{eq:dl}) is
solved using a coordinate-descent type algorithm, alternating between the de-aliasing
step and the data consistency step until convergence. In contrast, with CNNs, we are performing one step de-aliasing and the same network cannot be used to de-alias iteratively.
While CNNs may be powerful enough to learn one step reconstruction,
such a network could show signs of overfitting, unless there is vast
amounts of training data. In addition, training such networks may require a long
time as well as careful fine-tuning steps. It is therefore best to be able to
use CNNs for iterative reconstruction approaches.

\begin{figure*}[t]
  \centering
  \includegraphics[width=.9\textwidth]{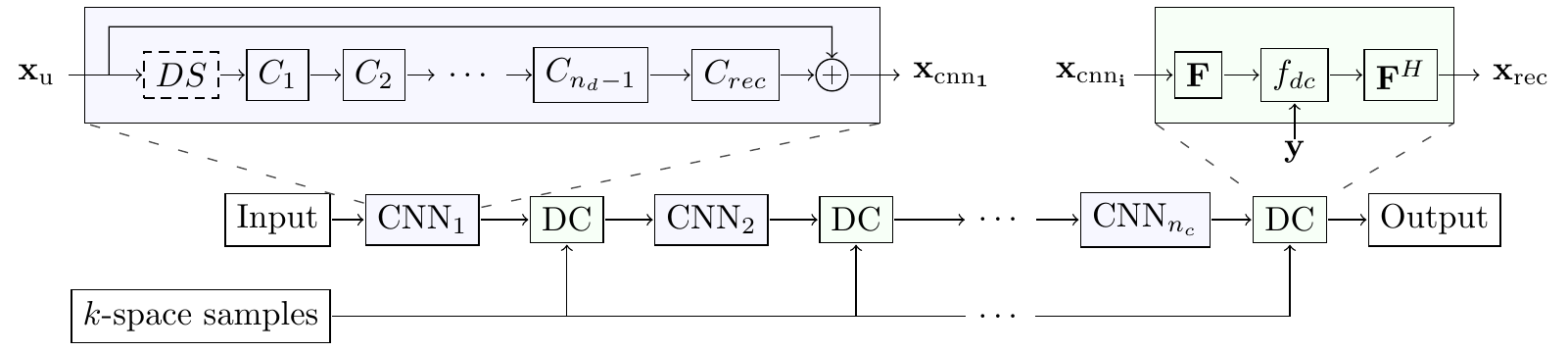}
  \caption{A cascade of CNNs. DC denotes the data consistency layer and DS denotes the data sharing layer. The number of convolution layers within each network and the depth of cascade is denoted by $n_d$ and $n_c$ respectively. Note also that DS layer only applies when the input is a sequence of images.}
\label{fig:custom_resnet_nd_nc} 
\end{figure*}

A simple solution is to train a second CNN which learns to reconstruct from the output of the first CNN. In fact, we can concatenate a new CNN on
the output of the previous CNN to build extremely deep networks which iterate
between intermediate de-aliasing and the data consistency reconstruction. We
term this a \emph{cascading network}. In fact, one can essentially
view this as unfolding the optimisation process of DLMRI. If each CNN expresses the dictionary learning reconstruction step, then the cascading CNN
can be seen as a direct extension of DLMRI, where the whole reconstruction
pipeline can be optimised from training, as seen in Fig. \ref{fig:custom_resnet_nd_nc}. In particular, owing to the forward and back-backpropagation rules defined for the DC layer, all subnetworks can be trained jointly in an end-to-end manner, defining yielding one large network.

%  \begin{figure*}[!t]
%   \centering
%   \includegraphics[width=1\textwidth]{kspace_avg_s}
%   \caption{Illustration of data sharing approach. (a) The ground truth image (b) 12-fold undersampling using the mask in Fig. \ref{fig:mask_along_t}a (c) Image generated by data sharing with $n_{adj} = 2$. The resulting image roughly appear as images acquired via 4-fold undersampling {\color{blue} (as opposed to theoretical maximum of 2.4-fold due to overlapped lines). The data sharing fills the entries indicated in white in Fig. \ref{fig:mask_along_t}b (d) Actual 4-fold undersampling of data by treating Fig. \ref{fig:mask_along_t}b as the undersampling mask } (e) The difference between (c) and (d). One can notice that the images are similar except for the data inconsistency of the dynamic content around heart.}
% \label{fig:kspace_avg} 
% \end{figure*}

% \begin{figure}[!t]
%   \centering
%   \includegraphics[width=.45\textwidth]{mask_along_t_labelled}
%   \caption{Cartesian undersampling mask displayed for $k$-$t$ axes, where sampled entries are indicated in light gray. The frequency encoding direction of $k$-space is fully sampled and hence omitted for simplicity.  (a) 12-fold undersampling mask (b) Entries that are filled using data sharing with $n_{adj}= 2$, conceptually simulating the 4-fold undersampling mask (c) Random 4-fold undersampling mask. Note that (c) achieves higher incoherence compared to (b) and therefore is preferred for the CS framework.}  
% \label{fig:mask_along_t} 
% \end{figure}

 \begin{figure}[t]
  \centering
  \includegraphics[width=.5\textwidth]{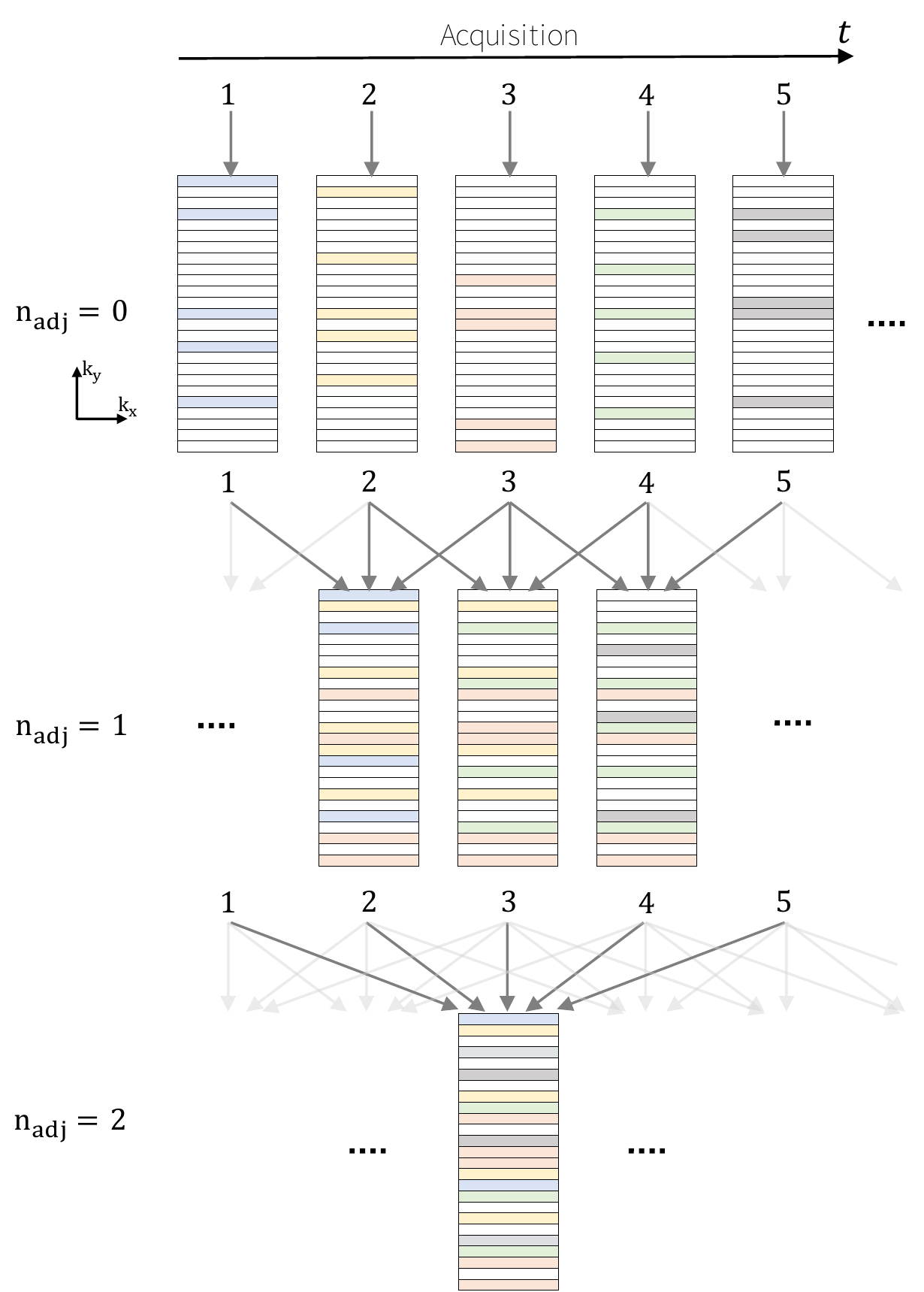}
  \caption{The illustration of data sharing approach. The acquired lines, which can be seen as $n_{adj}=0$, are colour-coded for each time frame. For each $n_{adj}$, the missing entries in each frame are aggregated using the values from up to $\pm n_{adj}$ neighbouring frames. The overlapped lines are averaged.}
\label{fig:kspace_avg1} 
\end{figure}

 \begin{figure}[t]
  \centering
  \includegraphics[width=.5\textwidth]{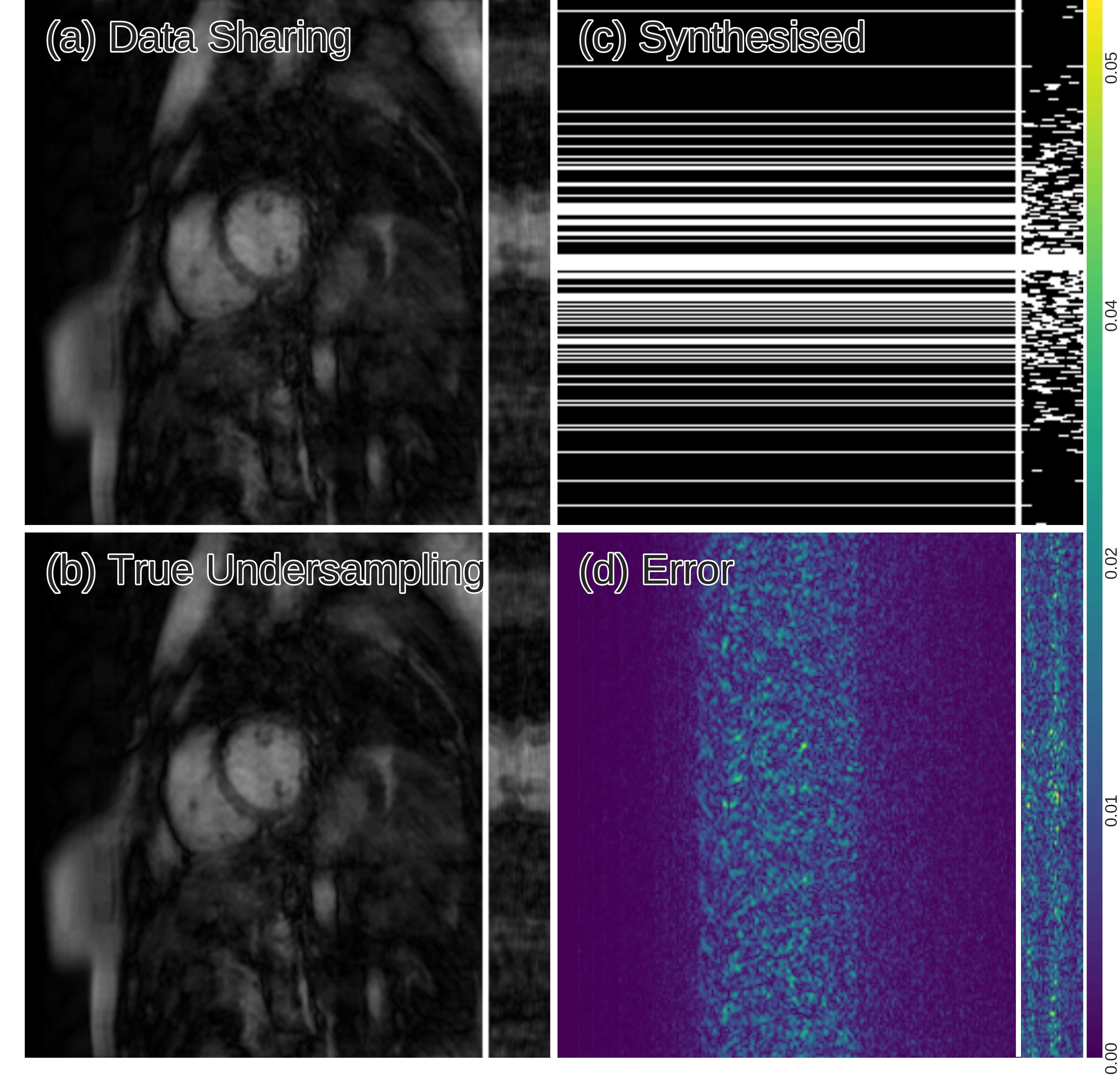}
  \caption{The illustration of data sharing approach applied to the image and the mask from Fig.\ref{fig:undersampling_example}(a,b). In this figure, (a) shows the appearance of the resulting sequence for $n_{adj} = 2$. (b) The entries in $k$-space that are either acquired or aggregated using the data sharing approach with $n_{adj} = 2$, which conceptually defines a sampling mask. (c) For a comparison, we show the resulting zero-filled reconstruction if (b) were treated as a mask. (d) The error map between the (a) and (b). One can observe their similarity except for the data \emph{inconsistency} of the dynamic content around the heart region. Note that for $n_{adj} = 2$, the obtained image has the appearance similar to acceleration factor around 4 (rather than 12/5 = 2.4, which is the maximum achievable from 5 frames) due to overlapping lines.}
\label{fig:kspace_avg2} 
\end{figure}

\section{Data Sharing Layer}

For the case of reconstructing dynamic sequences, the temporal correlation between frames can be exploited as an additional regulariser to further de-alias the undersampled images. For this, we use 3D convolution to learn spatio-temporal features of the input sequence. In addition, we propose incorporating features that could benefit the CNN reconstruction, inspired by \emph{data sharing} approaches \cite{riederer1988mr}, \cite{rasche1995continuous}, \cite{zhang2010magnetic}: if the change in image content is relatively small for any adjacent frames, then the neighbouring $k$-space samples along the temporal-axis often capture similar information. In fact, as long as this assumption is valid, for each frame, we can fill the entries using the samples from the adjacent frames to approximate missing $k$-space samples. Specifically, for each frame $t$, all frames from $t-n_{adj}$ to $t+n_{adj}$ are considered, filling the missing $k$-space samples at frame $t$. If more than one frame within the range contains a sample at the same location, we take the weighted average of the samples. The idea is demonstrated in Fig. \ref{fig:kspace_avg1}.

An example of data sharing with $n_{adj}=2$ applied to the Cartesian undersampling is shown in Fig. \ref{fig:kspace_avg2}(a). As data sharing aggregates the lines in $k$-space, the resulting images can be seen as a zero-filled reconstruction from a measurement with lower undersampling factor. In practice, however, cardiac sequences contain highly dynamic content around the heart and hence combining the adjacent frames results in data inconsistency around the dynamic region, as illustrated in Fig. \ref{fig:kspace_avg2}(b,c,d). However, for CNN reconstruction, we can incorporate these images as an extra input to train the network rather than treating them as the final reconstructions. Note that the reduction in the apparent acceleration factor is non-trivial to calculate: if each frame samples $10\%$ of $k$-space, combining 5 adjacent frames in theory should cover $50\%$. However, one often relies on variable density sampling, which samples low-frequency terms more often, yielding overlapped lines between the adjacent frames. Therefore, the apparent acceleration factor is often much less. As a remedy, regular sampling can be considered. However, regular sampling results in coherent artifact in the image domain, the removal of which is a different problem from the one we address here, which attempts to resolve \emph{incoherent} aliasing patterns. Alternatively, one can perform a sampling trajectory optimisation to reduce the overlapping factor, however, this is out-of-scope for this work and will be investigated in future. 

For our network, we implement \emph{data sharing (DS) layers} which take an input image and generate multiple ``data-shared" images for a range of $n_{adj}$. The resulting images are concatenated along the channel-axis and treated as a new input fed into the first convolution layer of the CNNs. Therefore, using the images obtained from data sharing can be interpreted as transforming the problem into joint estimation of aliasing as well as the dynamic motion, where the effect of aliasing is considerably smaller. Note that for the cascading network architecture, from the second subnetwork onwards, the input to each subnetwork is no longer "undersampled", but instead contains intermediate predicted values from the previous subnetwork. In this case, we average all the entries from the adjacent frames and update the samples which were not initially acquired. For this work, we allocate equal weight on all adjacent $k$-space samples, however, in future, more elaborate averaging schemes can be considered. 

\section{Architecture and Implementation}

Incorporating all the new elements mentioned above, we can devise our cascading network
architecture. Our CNN takes in a two-channeled sequence of images $\R^{2N_x N_y N_t}$, where the channels store real and imaginary parts of the zero-filled reconstruction in the image domain. Based on literature, we used the following network architecture for the CNN, illustrated in Fig. \ref{fig:custom_resnet_nd_nc}: it has $n_d-1$ 3D convolution layers $C_i$, which are all followed by Rectifier Linear Units (ReLU) as a choice of nonlinearity. For each of them, we used a kernel size $k = 3$ \cite{Szegedy2015} and the number of filters was set to $n_f = 64$. The final layer of the CNN module is a convolution layer $C_{\text{rec}}$ with $k=3$ and $n_f = 2$, which projects the extracted representation back to the image domain. We also used \emph{residual connection} \cite{He2015}, which sums the output of the CNN module with its input. Finally, we form a cascading network by using the DC layers interleaved with the CNN reconstruction modules $n_c$ times. For DS layer, we take the input to each subnetwork, generating images for all $n_{adj} \in \{0, 1, \dots, 5\}$. As aforementioned, the resulting images are concatenated along the channel-axis and fed to the first convolution layer. We found that this choice of architecture works sufficiently well, however, the parameters were not optimised and there is therefore room for refinement of the results presented. Hence the result is likely to be improved by, for example, incorporating pooling layers and varying the parameters such as kernel size and stride \cite{Ronneberger2015}, \cite{Yu2016}.

Our model can also be used for 2D image reconstruction by setting $N_t = 1$ and use 2D convolution layers instead, however, data sharing does not apply to 2D reconstruction. For the following experiments, we first explore the network configurations by considering 2D MR image reconstruction. We identify our network by the values of $n_c$, $n_d$ and the use of data sharing. For example, \emph{D5-C2} means a network with $n_d=5$, $n_c=2$ with no data sharing. \emph{D5-C10(S)} corresponds a network with $n_d=5$, $n_c=10$ and data sharing. 

As mentioned, pixel-wise squared error was used as the objective function. As the proposed architecture is memory-intensive, a small minibatch size is used to train the cascade networks. We used minibatch size 1 for all the experiments but we did not observe any problem with the convergence. We initialised the network weights using He initialisation \cite{He2015a}. The Adam optimiser \cite{Kingma2014} was used to train all models, with parameters $\alpha = 10^{-4}, \beta_1 = 0.9$ and $\beta_2 = 0.999$ unless specified. We also added $\ell_2$ weight decay of $10^{-7}$. 

\section{Experimental Results}
\subsection{Setup}

% \begin{figure}[!t]
% \centering
% \includegraphics[width=.15\textwidth]{experiment_target.png}\hfill
% \includegraphics[width=.15\textwidth]{experiment_mask.png}\hfill
% \includegraphics[width=.15\textwidth]{experiment_input.png}
% \caption{(Left) Ground truth (Middle) Cartesian undersampling mask for
%   4-fold acceleration 
%   and (Right) A zero-filled reconstruction.}
% \label{fig:4x_und}
% \end{figure}

\paragraph{Dataset}

Our method was evaluated using the cardiac MR dataset consisting of 10 fully sampled short-axis cardiac cine MR scans. Each scan contains a single slice SSFP acquisition with 30 temporal frames with a $320 \times 320$ mm field of view and 10 mm slice thickness. The data consists of 32-channel data with sampling matrix size $192\times 190$, which was zero-filled to the matrix size $256\times256$. The raw multi-coil data was reconstructed using SENSE \cite{Pruessmann1999} with no undersampling and retrospective gating. Coil sensitivity maps were normalized to a body coil image to produce a single complex-valued image set that could be back-transformed to regenerate complex $k$-space samples or further processed to form final magnitude images. For the following experiments, we perform retrospective undersampling, simulating a practical single-coil acquisition scenario.

 \begin{figure}[t]
  \centering
  \includegraphics[width=.4\textwidth]{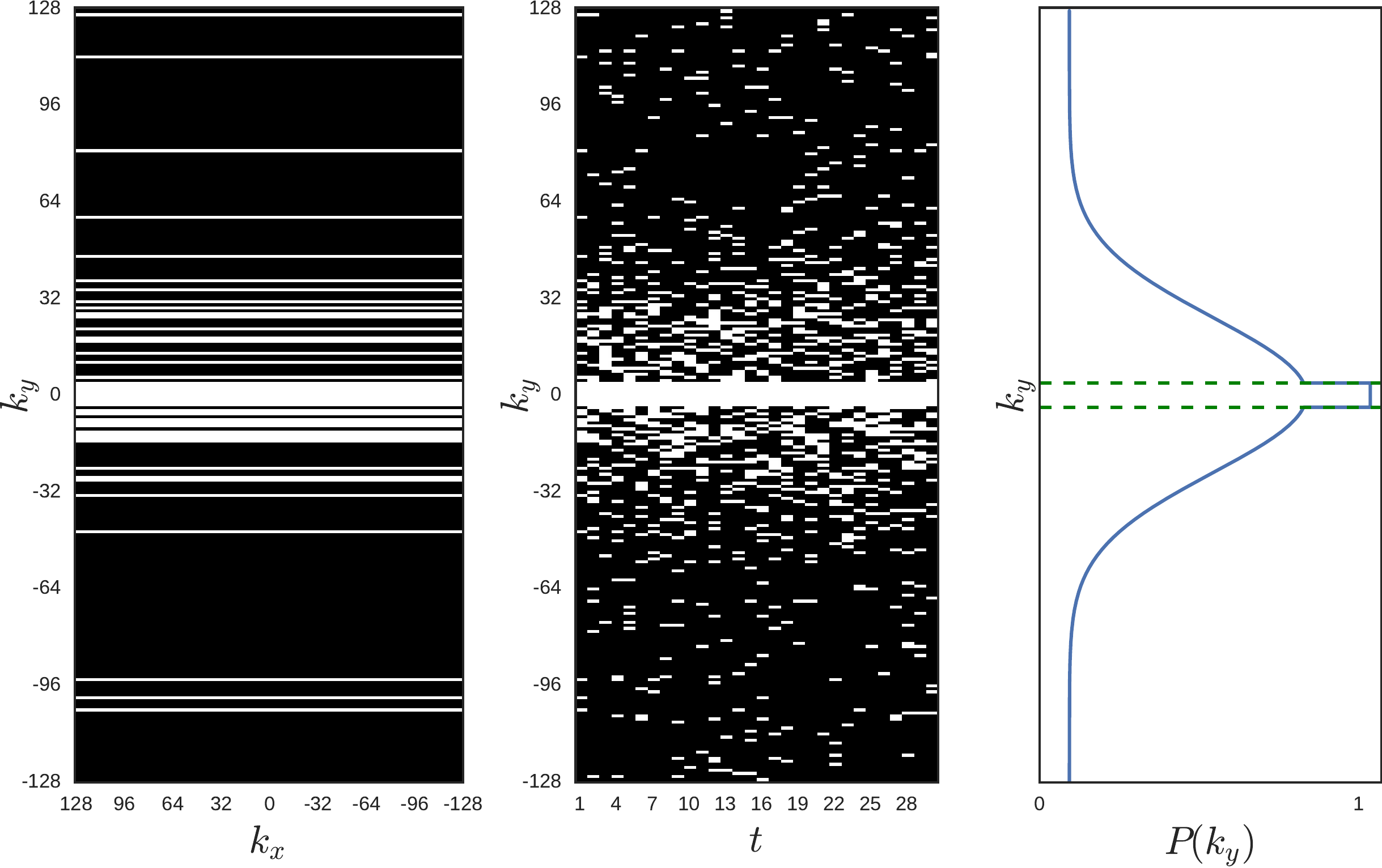}
  \caption{The detail of the Cartesian undersampling mask employed in this work. Note that the mask can be seen as a 3D volume indexed by $(k_x,k_y,t)$. For each image frame $t$, we fully sample along $k_x$-axis and undersample in $k_y$ direction. We always acquire the 8 central lines and the remaining lines are sampled according to a zero-mean Gaussian distribution with the tail that is marginally offset so it will never reach zero.}
\label{fig:undersampling_mask} 
\end{figure}

\paragraph{Undersampling} In this work, we focus on Cartesian undersampling, where one fully samples frequency-encodes (along $k_x$) and randomly undersamples the phase encodes (along $k_y$). In addition, we pair consecutive phase encodes, which has been reported to reduce eddy current which is a source of image degredation \cite{Markl2005}. For each frame, the eight lowest spatial frequencies are always acquired and other frequencies have a probability of being acquired determined by a zero-mean Gaussian variable density function that is marginally offset, such that the probability of acquisition never reaches zero even at the highest frequencies. An implementation of this approach can be found in \cite{Jung2007}, and an example of a 2D mask and its effect on the magnitude of a temporal frame is shown in Fig. \ref{fig:undersampling_mask}. For each experiment, the undersampling rate is fixed and will be stated. For training, the sampling masks were generated on-the-fly to allow the network to learn the differences between potential aliasing artefacts and the underlying signal better. Note that for each acceleration factor $acc$, one can generate ${{k_y}\choose{k_y / acc}}$ different masks.
     
While Cartesian acquisition is the most common protocol in practice and offers straightforward implementation using fast Fourier transform (FFT), other practical sampling strategies such as radial \cite{Block2007} or spiral \cite{Lustig2005} could be considered, which achieve greater aliasing incoherence. Nevertheless, they require the use of methods such as nonuniform Fourier transforms and gridding \cite{Fessler2007} which could propagate interpolation errors. 
%      In addition, greater aliasing incoherence can be achieved with 2D-space undersampling [cite], frequency encodes can be considered instantaneous relative to phase encodes, so acceleration is only meaningful through phase encode undersampling.

\paragraph{Data Augmentation} Typically, deep learning benefits from large datasets, which are often not available for medical images. Our dataset is relatively small (300 images), however, the literature suggests that it is still possible to train a network
% . In \cite{Dong2016}, the authors trained a network for super-resolution task on patches of 90 images, which showed comparable performance to the network trained on ImageNet, a dataset containing more than one million images. Similarly, \cite{Ronneberger2015} showed that 
by applying appropriate data augmentation strategies \cite{Ronneberger2015}.
% , it can train a large network for segmentation task from only 30 images.
Therefore, we follow that practice and apply data augmentation including rigid transformation and elastic deformation to counter overfitting. Specifically, given each image (or a sequence of images), we randomly apply translation up to $\pm 20$ pixels along $x$ and $y$-axes, rotation of $[0, 2\pi)$, reflection along $x$-axis by $50\%$ of chance. Therefore, from rigid transformation alone, we create $0.3$ million augmented data per image. Combined with the on-the-fly generation of undersampling masks, we generate very large dataset. For the dynamic scenario, we further added elastic deformation, using the implementation in \cite{simard2003best}, with parameters $\alpha \in [0, 3]$ and $\sigma \in [0.05, 0.1]$, sampled uniformly, as well as reflection along temporal axis. Note that while strong elastic deformation may produce anatomically unrealistic shapes its use is justified as our goal is to train a network which learns to \emph{de-alias} the underlying object in the image, rather than explicitly learning the anatomical shapes.

% {\color{blue} In addition to data augmentation for 2D, we also add flipping along temporal dimension. while this is unrealistic, it is justified for the purpose of training, as we are interested in learning a denoiser and not an explicit the motion model. If we have enough training data, then this is plausible and potentially further improve the reconstruction, but this (1) may not generalise to patients with irregular heartbeat (2) it would become necessary to learn the entire volume. We further add elastic deformation to give, practically, infinitely many variations. 
% }

\paragraph{Evaluation Methodology}

For the 2D experiments, we split the dataset into training and testing sets including 5 subjects each. Each image frame in the sequence is treated as an individual image, yielding a total of 150 images per set. Note that typically, a portion of training data is treated as a validation set utilised for early-stopping \cite{Bishop2006}, where one halts training if the validation error starts to increase. Initially, we used 3-2-5 split for training, validation and testing. However, even after 3 days of training \emph{cascade} networks, we did not observe any decrease in the validation error. Therefore, we instead included the validation set in the training to further improve the performance but fix the number of backpropagation to be an order of $10^5$, which we empirically found to be sufficient. For the dynamic experiments, we used 7-3 split for training and testing and an order of $10^4$ for the number of backpropagation.  

To evaluate the performances of the trained networks, we used mean squared error (MSE) as our quantitative measure. The reconstruction signal-to-noise ratio from undersampled data is highly dependent on the imaging data and the undersampling mask. To take this into consideration for fair comparison, we assigned an arbitrary but fixed undersampling mask for each image in the test data, yielding a fixed number of image-mask pairs to be evaluated.

\subsection{Reconstruction of 2D Images}

\subsubsection{Trade-offs between $n_d$ and $n_c$}
\label{sec:ex1}

In this experiment we compared two architectures: \emph{D5-C2}  ($n_d=5,n_c=2$) and \emph{D11-C1} ($n_d = 11, n_c= 1$) to evaluate the benefit of the DC step. The two networks have equivalent depths when the DC layers are viewed as feature extraction layers. However, the former can build deeper features of the image, whereas the latter benefits from the intermediate data consistency step. The undersampling rate was fixed to 3-fold and each network was trained end-to-end for $3 \times 10^5$ backpropagations. 

\begin{figure}[!t]
  \centering
%   \centerline{\includegraphics[width=0.5\textwidth]{ex5_2_2}}
%   \centerline{\includegraphics[width=0.4\textwidth]{paper_ex1_5_5_compact}}
  \centerline{\includegraphics[width=0.4\textwidth]{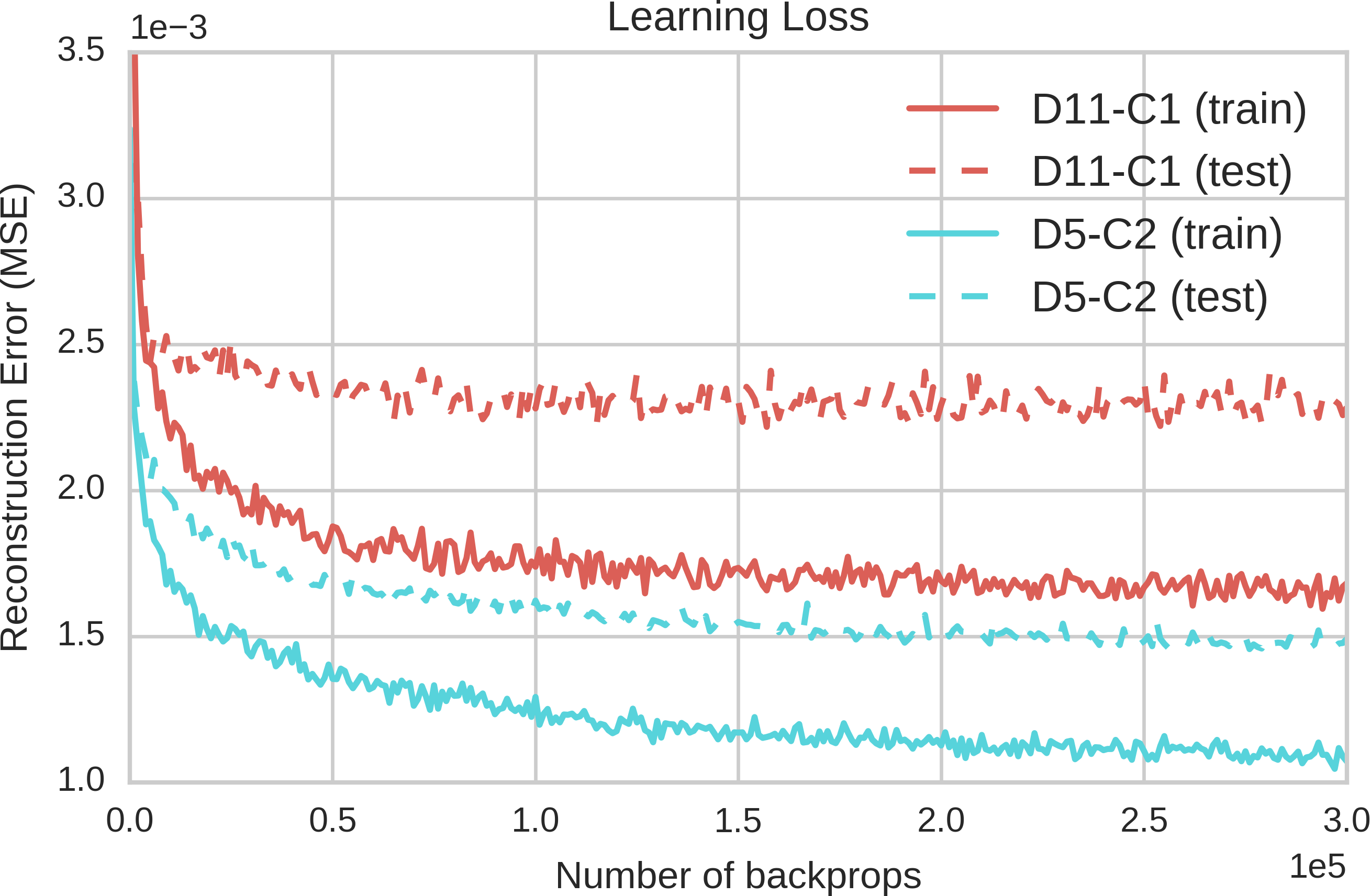}}
  \caption{A comparison of  the networks with and without the intermediate DC step. \emph{D5-C2} shows superior performance over \emph{D11-C1}. In particular, \emph{D5-C2} has considerably lower test error, showing an improved generalization property.}
\label{fig:ex1} 
\end{figure}

The MSE's on the training and test data are shown in Fig. \ref{fig:ex1}. Note that a gap between the performance on training and test set may exist by the nature of the dataset (e.g. due to image features, initial level of aliasing, etc.) and therefore it is more informative to study in combination the rate of improvement and the slope at the tail of the curves to assess the overfitting process. Indeed, one can observe that \emph{D11-C1} eventually started to overfit the training data after about $1.2\times 10^5$ backpropagations. As one would expect, since our dataset is small, deep networks can overfit easily. On the other hand, both train and test errors for \emph{D5-C2} were notably lower and had relatively tighter gap, showing better generalisability compared to \emph{D11-C1}. This is suggestively because the architecture employs two data consistency steps and rebuilds the representations at each cascading iteration. This suggests that it is more beneficial to interleave DC layers projecting the acquired $k$-space onto intermediate reconstructions with the CNN image reconstruction modules, which appears to help both the reconstruction as well as the generalisation. Nevertheless, there is a considerable gap between train and test data even for \emph{D5-C2}. 
% While it is fairly common for many learning algorithm to have much lower training error, 
However, we note from the figure that even after $3 \times 10^5$ backpropagations, the test error is still improving. Therefore, although it seems that the network gets more optimised to the features in training data quickly, it still learns features generalisable to test data. Having more training data is likely to accelerate the learning process.

\subsubsection{Effect of Cascading Iterations $n_c$}
\label{sec:ex3.1}

In this experiment, we explored how much benefit the network can get by increasing the cascading iteration. We fixed the architectures to have $n_d=5$, but varied the cascading iteration $n_c \in \{1, 2, 3, 4, 5\}$. For this section, due to time constraints, we trained the networks using a greedy approach: we initialised the cascading net with $n_c = k$ using the net with $n_c=k-1$ that was already trained. For each $n_c$, we performed $10^5$ backpropagations. Note that the greedy approach leads to a satisfactory solution, however, better results can be achieved with random initialisation, as initialising a network from another networks convergence point can make it more likely that it gets stuck in suboptimal local minima.

\begin{figure}[!t]
  \centering
%   \centerline{\includegraphics[width=0.5\textwidth]{ex5_2_2}}
  \centerline{\includegraphics[width=0.45\textwidth]{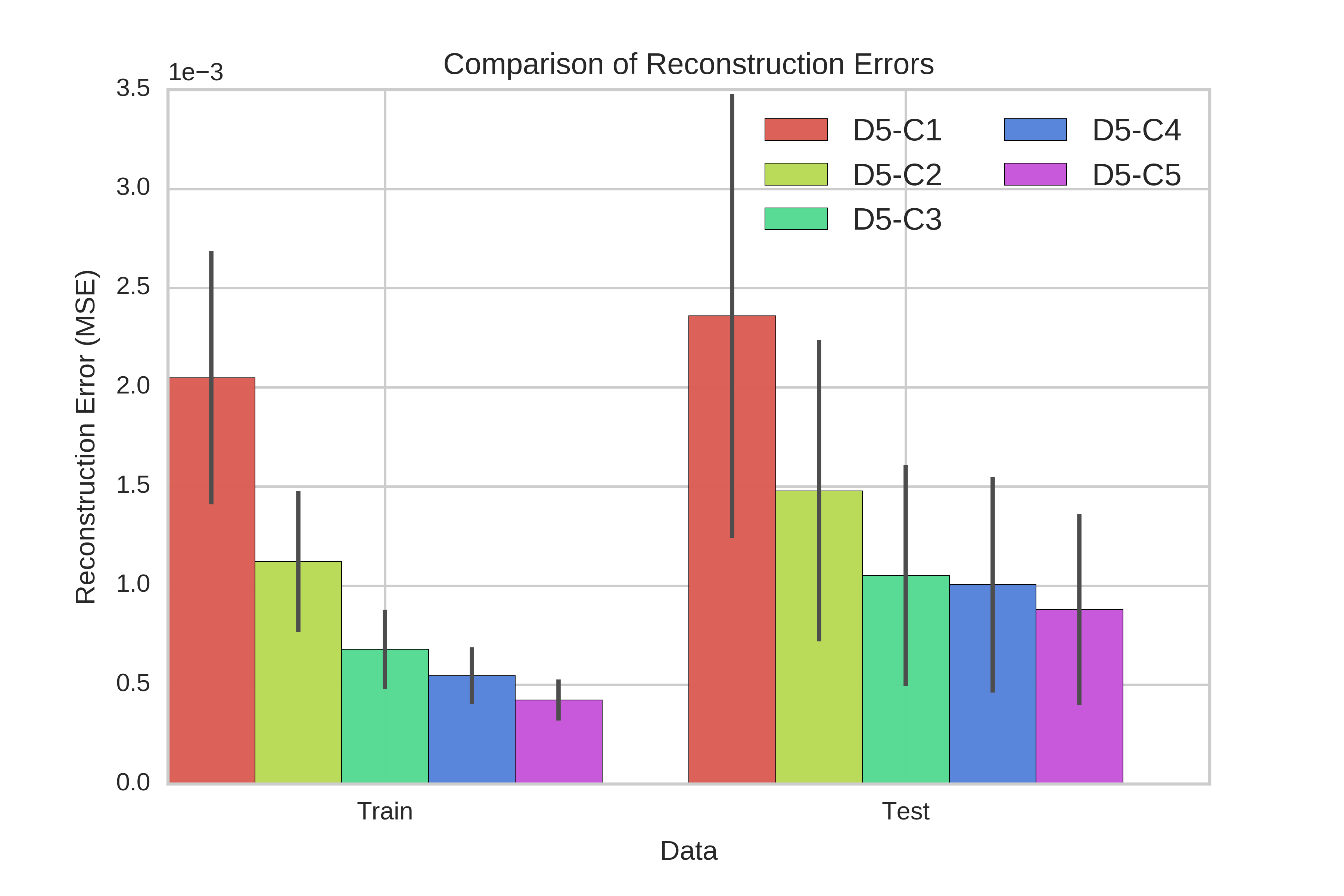}}
  \caption{The effect of increasing cascading iteration $n_c$. One can see that the reconstruction error on both training and test data monotonically decreases as $n_c$ increases. However, the rate of  improvement is reduced after $n_c=3$.}
\label{fig:greedy_layer_error_bars} 
\end{figure}

Reconstruction errors for each cascading network of different $n_c$ are shown in Fig. \ref{fig:greedy_layer_error_bars}. We observed that while deeper cascading nets tend to overfit more, they still reduced the test error every time. The rate of improvement was reduced after 3 cascading layers, however, we see that the standard deviation of error was also reduced for the deeper models. In the interest of space, we have not shown the resulting images of each \emph{D5-C$n_c$} but we have observed the that increasing $n_c$ resulted in images with more of the subtle image details correctly reconstructed and there were also less noise-like aliasing remaining in the images.

\begin{figure}[!t]
  \centering
%   \centerline{\includegraphics[width=0.49\textwidth]{paper_ex4}}
  \centerline{\includegraphics[width=.5\textwidth]{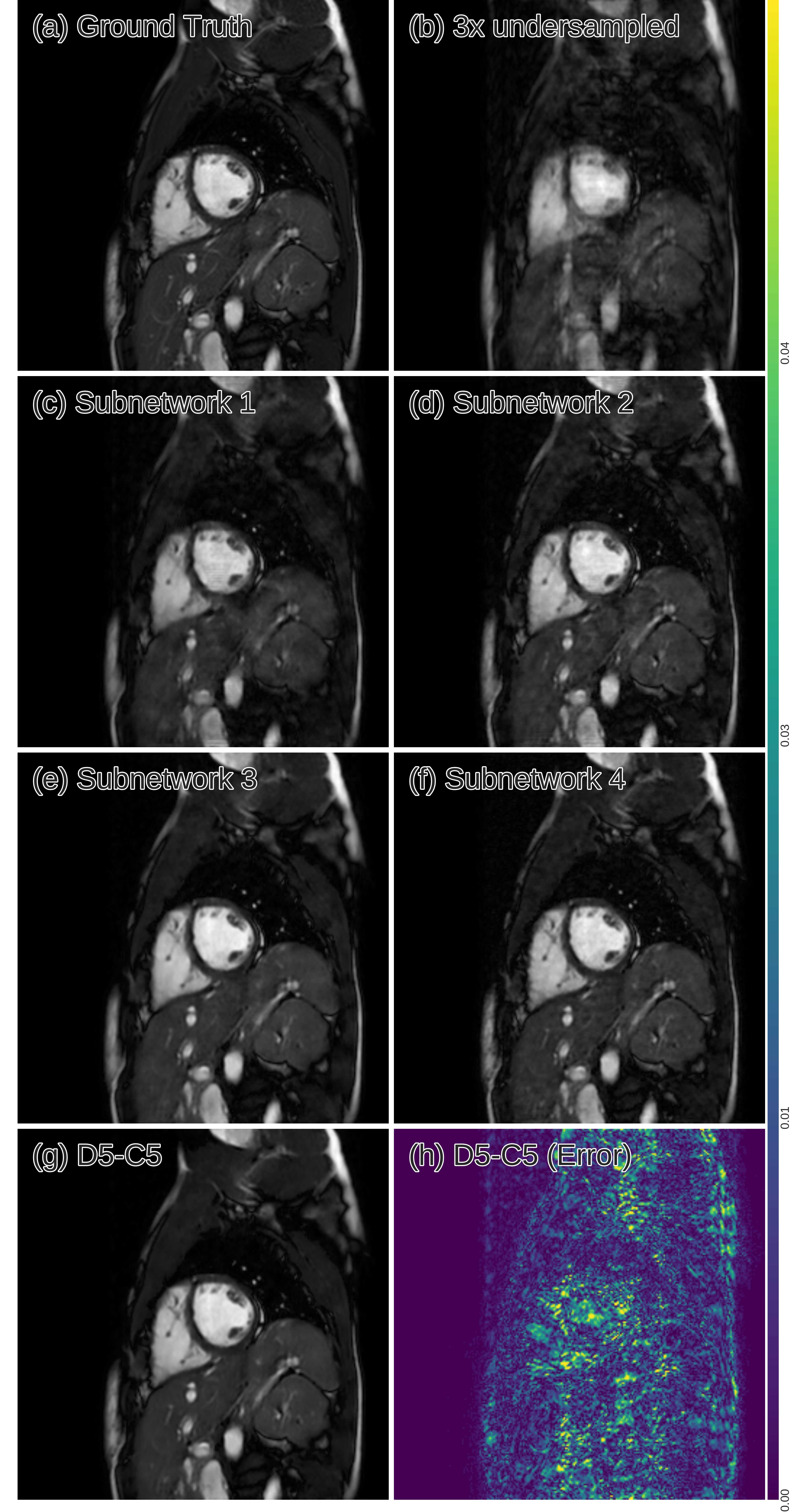}}
  \caption{2D reconstruction results of $D5$-$C5$ for one of the test subjects. Here we inspect 
  the intermediate output from 
    each 
    subnetwork in \emph{D5-C5}. (a) Ground truth (b) The input to the network that was 3x
    undersampled image. The output of (c) first, (d) second, (e) third, (f) fourth
    cascading subnetwork respectively. (g,h) The final output and the
    corresponding error. Note that this is not the reconstruction results from the networks in Experiment in \ref{sec:ex3.1}.}
\label{fig:c5_depth}
\end{figure}

On the other hand, in Fig. \ref{fig:c5_depth}, we show the \emph{intermediate} reconstructions from each subnetwork within \emph{D5-C5} to better understand how the network exploits the iterative nature internally. In general, we see that the cascading
net gradually recovers and sharpens the output image. Although the reconstruction error decreased monotonically at each cascading depth, we observed that the output of the fourth subnetwork appears to be more grainy than the output of the preceding subnetwork. This suggests the benefit of the end-to-end training scheme: since we are optimising the whole pipeline of reconstruction, the additional CNN’s are internally used to rectify the error caused by the previous CNN’s. In this case, the fourth subnetwork appears to counteract  over-smoothing in the third subnetwork. 

\subsubsection{Comparison with DLMRI}
\label{sec:vs_dl_2d}

In this experiment, we compared our model with the state-of-the-art DL-based method, DLMRI, for reconstructing individual 2D cardiac MR images. The comparison was performed for 3-fold and 6-fold acceleration factors. 

\paragraph{Models} 

For CNN, we selected the parameters $n_d = 5$, $n_c=5$. To ensure a fair comparison, we report the aggregated result on the test set from two-way cross-validation (i.e. two iterations of train on five subjects and test on the other five). For each iteration of the cross validation, the network was end-to-end trained using He intialisation \cite{He2015a}. For 6-fold undersampling, we initialised the network using the parameters obtained from the trained models from 3-fold acceleration. Each network was trained for $3 \times 10^5$ backpropagations, which took one week to train per network on a GeForce TITAN X, however, our manual inspection of the loss curve indicates that the training error plateaued at much early stage, approximately within 3 days.

 For DLMRI, we used the implementation from \cite{Ravishankar2011} with patch size $6 \times 6$. Since DLMRI is quite time consuming, in order to obtain the results within a reasonable amount of time, we trained a joint dictionary for all time frames within the subject and reconstructed them in parallel. Note that we did not observe any decrease in performance from this approach. For each subject, we ran 400 iterations and obtained the final reconstruction.

\paragraph{Results}
\label{sec:vs_dl_2d_res}

\begin{table}[t]
\renewcommand{\arraystretch}{1.3}
\caption{The result of 2D reconstruction. DLMRI vs. CNN across 10 scans}
\label{table:vs_dl}
\centering
\begin{tabular}{c|c||c|}
\cline{2-3}
  & \multicolumn{1}{c||}{3-fold} & \multicolumn{1}{c|}{6-fold} \\
\hline
\multicolumn{1}{|c|}{Models} & MSE (SD) $\times 10^{-3}$ & MSE (SD) $\times 10^{-3}$ \\
\hline
\multicolumn{1}{|c|}{DLMRI} & 2.12 (1.27) & 6.31 (2.95) \\
\hline
\multicolumn{1}{|c|}{CNN (2D)} & \textbf{0.89} (\textbf{0.46}) & \textbf{3.42} (\textbf{1.65}) \\
\hline
\end{tabular}
\end{table}

% \begin{figure}[!t]
%   \centering
% %   \includegraphics[width=0.45\textwidth]{paper_ex5}
%   \includegraphics[width=\textwidth]{paper_ex5_side}
%   \caption{The comparison of reconstruction from DLMRI and CNN. (a) The original,
%     (b) 3x undersampled, (c)-(d) DLMRI reconstruction and its error map $\times
%     5$ and (e)-(f) CNN reconstruction and its error map $\times 5$. }
%   \label{fig:tmi2_ac4_detailed}
% \end{figure}

\begin{figure}[!t]
  \centering
  \includegraphics[width=.5\textwidth]{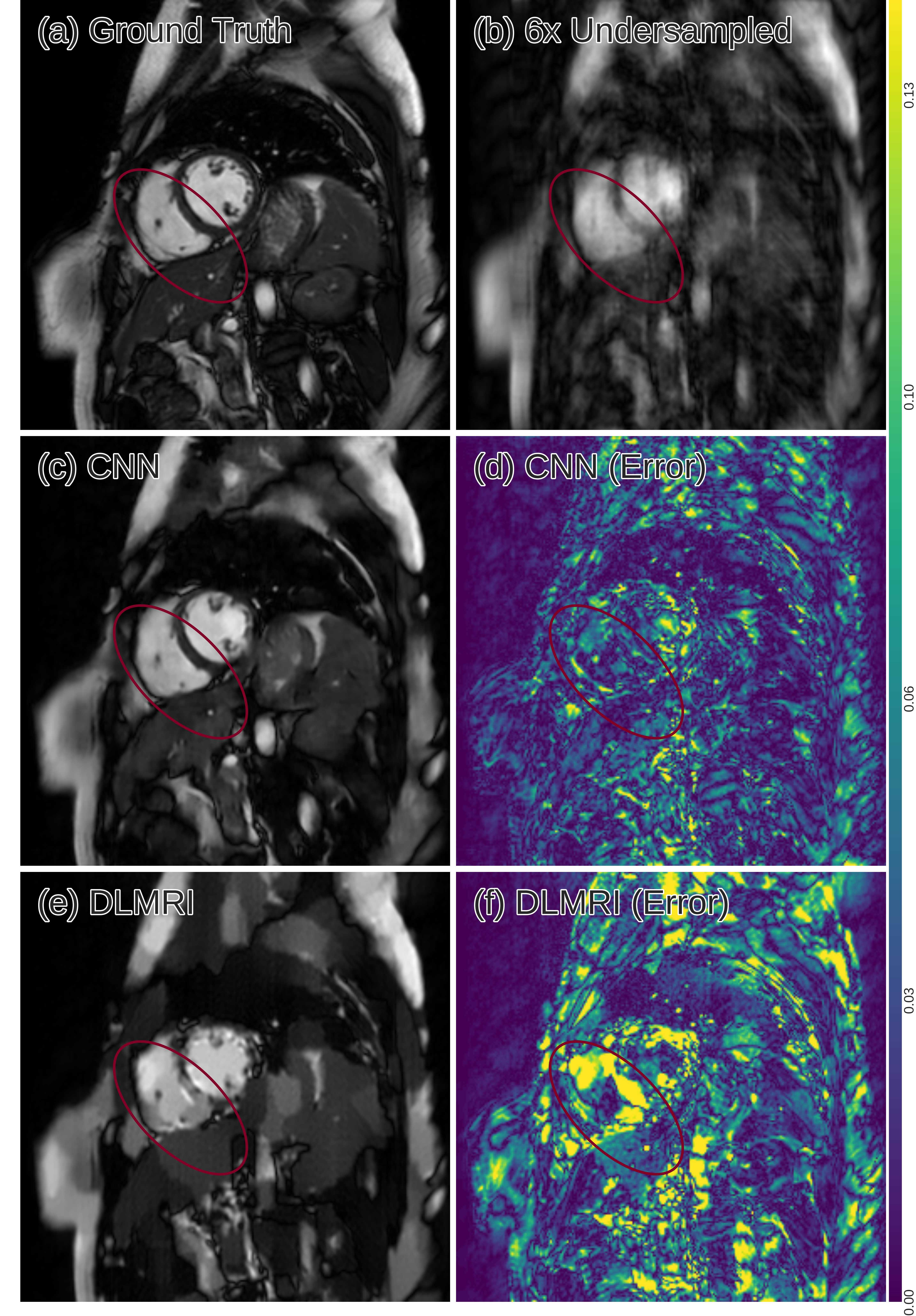}
  \caption{The comparison of 2D reconstructions from DLMRI and CNN for test data. (a) The original
    (b) 6x undersampled (c,d) CNN reconstruction and its error map
    (e,f) DLMRI reconstruction and its error map. There are larger errors in (f) than (d) and red ellipse highlights the anatomy that was reconstructed by CNN better than DLMRI.}
\label{fig:tmi2_ac8_detailed}
\end{figure}

% \paragraph{Quantitative Result} 
% \st{We recorded the MSE for each reconstructed slice, however, to report them concisely, we aggregated the results of all subjects, reporting their means.} 
The means of the reconstruction errors across 10 subjects are summarised in Table. \ref{table:vs_dl}. For both 3-fold and 6-fold acceleration, one can see that CNN consistently outperformed DLMRI, and that the standard deviation of the error made by CNN was smaller. The reconstructions from 6-fold acceleration is in Fig. \ref{fig:tmi2_ac8_detailed}. Although both methods suffered from significant loss of structures, the CNN was still capable of better preserving the texture than DLMRI (highlighted in red ellipse). On the other hand, DLMRI created block-like artefacts due to over-smoothing. 6x undersampling for these images typically approaches the limit of sparsity-based methods, however, the CNN was able to predict some anatomical details which was not possible by DLMRI. This could be due to the fact that the CNNs has more free parameters to tune with, allowing the network to learn complex but more accurate end-to-end transformations of data.  

\paragraph{Comparison of Reconstruction Speed} While training the CNN is time consuming, once it is trained, the inference
can be done extremely quickly on a GPU. Reconstructing each slice took  $23 \pm 0.1$
milliseconds on a GeForce GTX 1080, which enables real-time applications. To produce the above results, DLMRI took
about $6.1 \pm 1.3$ hours per subject on CPU. Even though we do not have a GPU implementation of DLMRI, it is expected to take longer than 23ms because DLMRI requires dozens of iterations of dictionary learning and sparse coding steps. Using a fixed, pre-trained dictionary could remove this bottleneck in computation although this would likely be to the detriment of reconstruction quality. 
% (hence $690\pm3$ milliseconds for the whole subject) 
\subsection{3D Experiments}

For the following experiments, we split our dataset into training and testing sets containing seven and three subjects respectively. Compared to the 2D case, we have significantly less data. As aforementioned, we applied elastic deformations in addition to rigid transformation to augment the training data input in order to increase the variation of the examples seen by the network. Furthermore, working with a large input is a burden on memory, limiting the size of the network that can be used. To address this, we trained our model on an input size $256\times N_{patch} \times 30$, where the direction of patch extraction corresponds to the frequency-encoding direction. In this way, we can train the network with the same aliasing patterns while reducing the input size. Note that the extracted patches of an image sequence will have different $k$-space values compared to the original data once the field-of-view (FOV) is reduced. As such, this trick only works for training where the patches can be treated as the new instances of training data. In particular at test time, since only the raw data with full FOV is available, the CNN must also be applied to the entire volume in order to perform data consistency step correctly.

\subsubsection{Effect of Data Sharing}
\label{sec:3d_ex1}

\begin{figure}[!t]
  \centering
  \includegraphics[width=.4\textwidth]{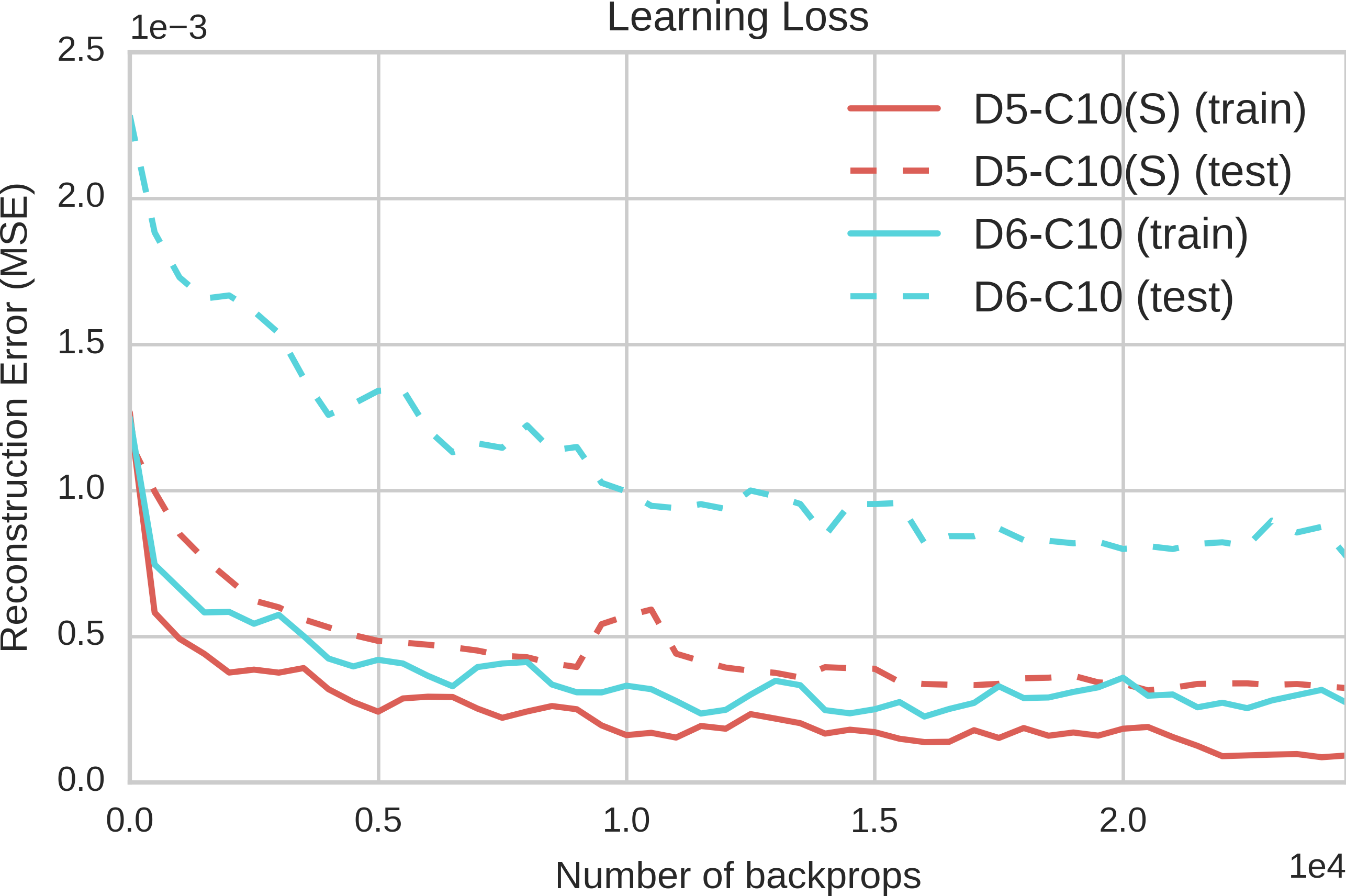}
  \caption{The effect of data sharing. The network with data sharing shows superior performance over the other. In particular, it has considerably lower test error, showing an improved generalization property.}
  \label{fig:3d_ex1}
\end{figure}

In this experiment, we evaluated the effect of using the features obtained from data sharing. We trained the following two networks: \emph{D5-C10(S)} ($n_d=5$, $n_c=10$ with data sharing) and \emph{D6-C10} ($n_d=6$, $n_c=10$ without data sharing). In the second network, the data sharing is replaced by an additional convolution layer to account for the additional input. We trained each model to reconstruct the sequences from 9-fold undersampling for $2.5 \times 10^4$ backpropagations. Their learning is plotted in Fig. \ref{fig:3d_ex1}. We can notice that there is a considerable difference in their errors. The error of the \emph{D5-C10(S)} was smaller for both train and test, suggesting that it was able to learn a strategy to de-alias image that generalises better. Moreover, by using data sharing, the network was able to learn faster. The visualization of their reconstructions can be found in the following section.

\subsubsection{Comparison with State-of-the-art}
\label{sec:vs_dl_3D}

In this experiment, we compared our model with state-of-the-art methods: DLTG  \cite{Caballero2014}, kt-SLR \cite{lingala2011} and L+S \cite{otazo2014} for reconstructing the dynamic sequence. We compared the results for 3, 6, 9 and 11-fold acceleration factors.

\paragraph{Models} For the CNN, we used $n_d = 5$, $n_c = 10$ with data sharing as explained above. We also set the weight decay to 0 as we did not notice any overfitting of the model. Contrary to the 2D case, we trained each network as follows: we first pre-trained the network on various undersampling rates (0-9x) for $5 \times 10^4$ backpropagations. Subsequently, each network was fine-tuned for a specific undersampling rate using Adam with learning rate reduced to $5 \times 10^{-5}$ for $10^4$ backpropagations. We performed three way cross validation (where for two iterations we train on 7 subjects then test on 3 subjects, one iteration where we train on 6 subjects and test on 4 subjects) and we aggregated the test errors. The pre-training and the fine tuning stages took approximately 3.5 days and 14 hours respectively using a GeForce GTX 1080. Since the training is time consuming, we did not train the networks longer but we speculate that the network will benefit from further training using lower learning rates. For DLTG, we used the default parameters described in \cite{Caballero2014}. For kt-SLR, we performed grid search to identify the optimal parameters for the data, which were $\mu_1 = 10^{-5}$, $\mu_2 = 10^{-8}$, $\rho = 0.1$. Similarly for L+S, the optimal parameters were $\lambda_L=0.01$ $\lambda_S=0.01$. 

\paragraph{Result}

\begin{figure}[t]
  \centering
  \includegraphics[width=.4\textwidth]{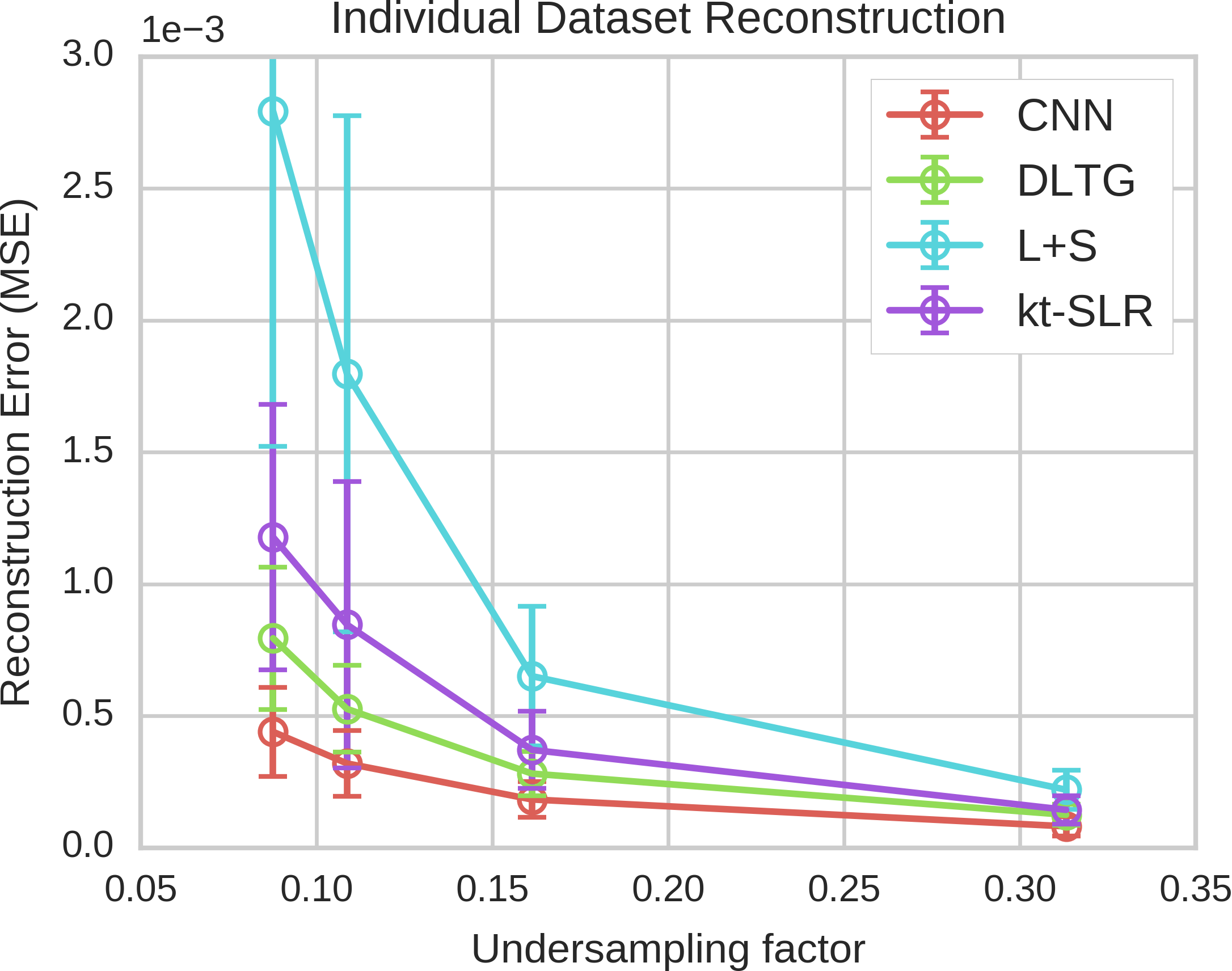}
  \caption{The reconstruction errors of CNN vs state-of-the-art methods across 10 subjects for different undersampling rates. Note that we average over the test error from all iterations of cross-validation.} 
\label{fig:vs_dltg_plot}
\end{figure}

\begin{figure}[t]
  \centering
  \includegraphics[width=.5\textwidth]{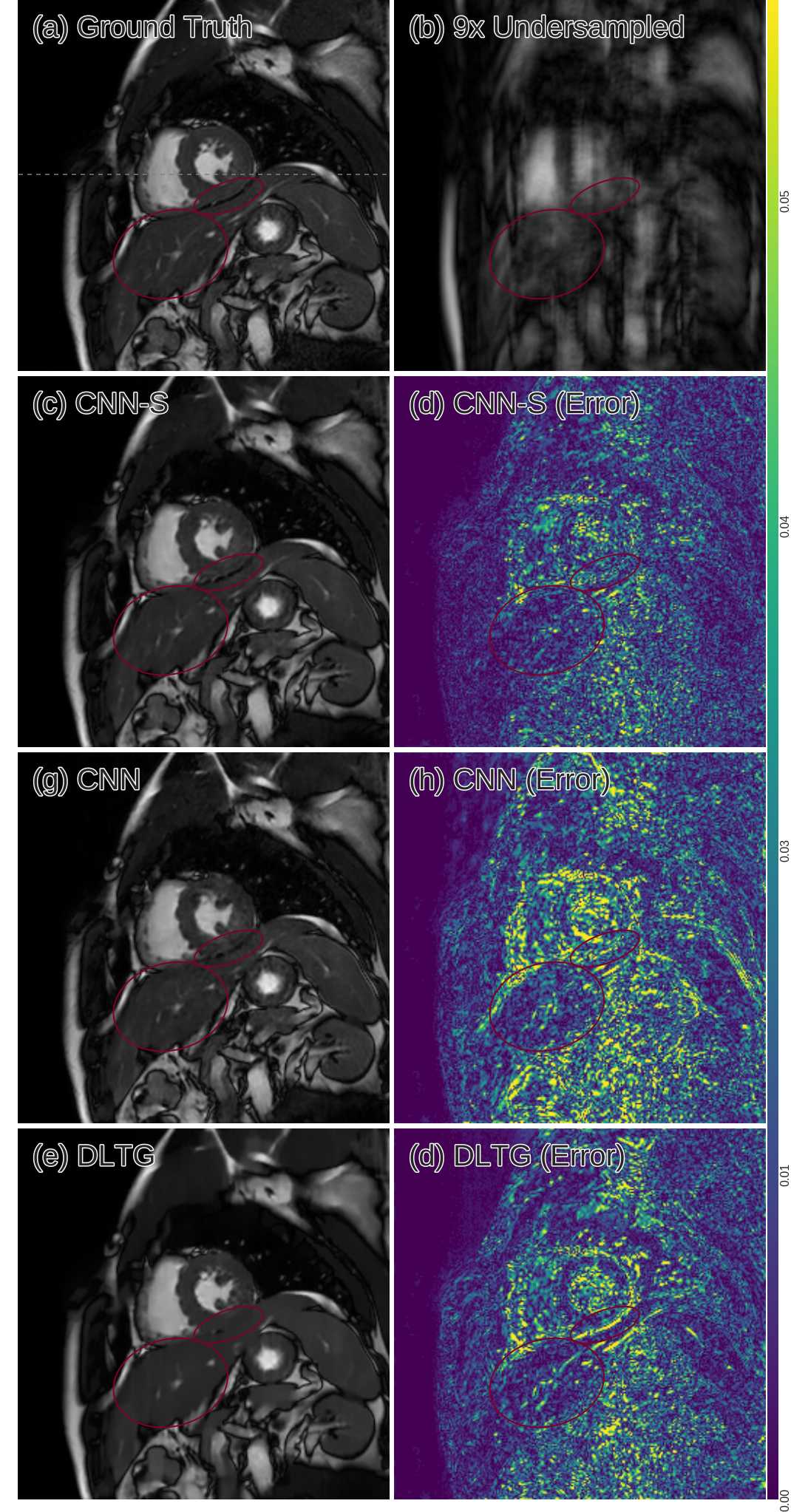}
  \caption{The comparison of cardiac MR image sequence reconstructions from DLTG and CNN. Here we show $n$th slice from one of the test subjects (a) The original
    (b) 9x undersampled (c,d) CNN with data sharing and its error map 
    (e,f) CNN without data sharing and its error map 
    (g,h) DLTG reconstruction and its error map. Red ellipses highlight the anatomy that was reconstructed by CNN better than DLTG.}
\label{fig:vs_dltg1}
\end{figure}

\begin{figure}[t]
  \centering
  \includegraphics[width=0.4\textwidth]{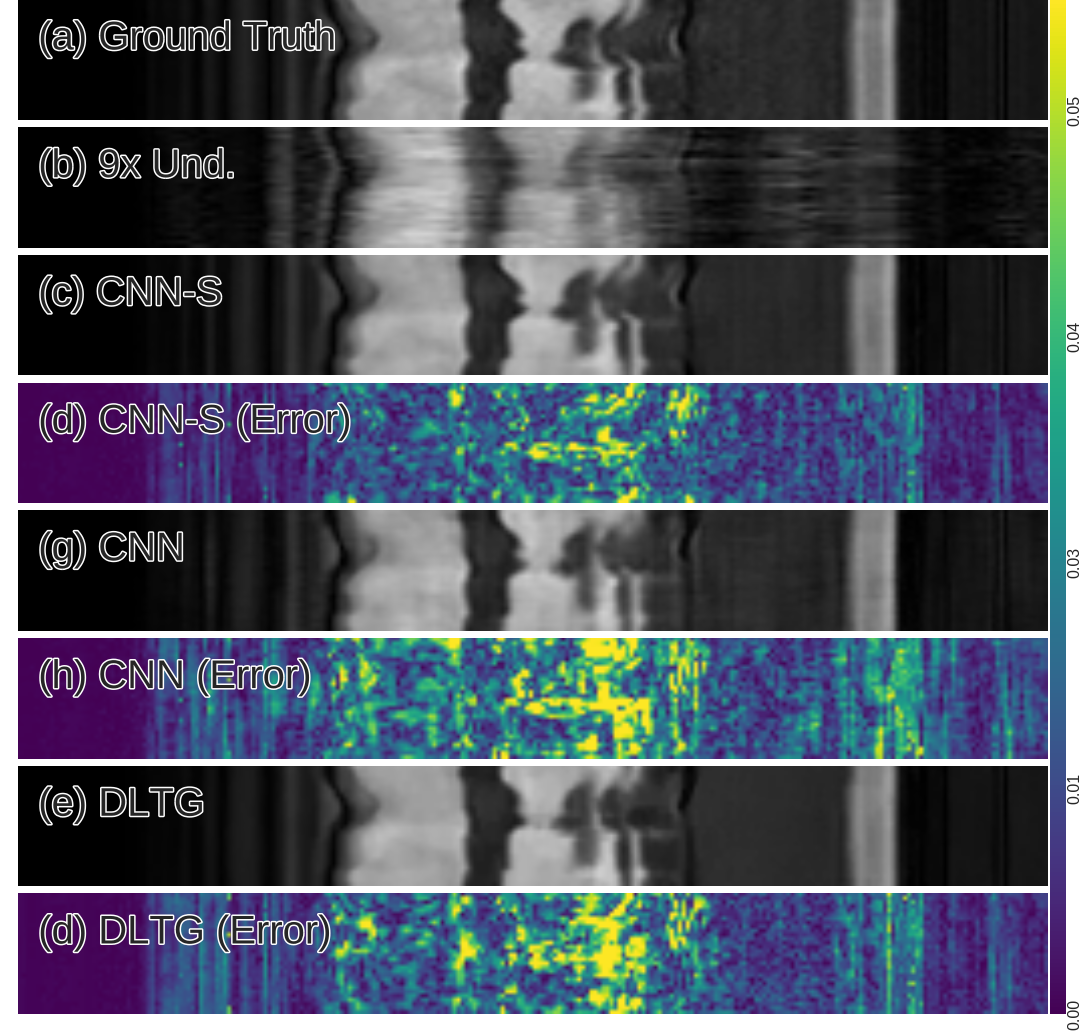}
  \caption{The comparison of reconstructions along temporal dimension. Here we extract a 110th slice along y-axis from the previous figure. (a) The original
    (b) 9x undersampled (c,d) CNN with data sharing and its error map     (e,f) CNN without data sharing and its error map 
    (g,h)  DLTG reconstruction and its error map. 
    }
\label{fig:vs_dltg2}
\end{figure}

The final reconstruction error is summarised in Fig. \ref{fig:vs_dltg_plot}. we see that CNN consistently outperforms state-of-the-art methods for all undersampling factors. For a low acceleration factor (3x undersampling), all methods performed approximately the same, however, for more aggressive undersampling factors, CNN was able to reduce the error by a considerable margin. For aggressive undersampling rates, the performance of kt-SLR and L+S degraded much faster. These methods employ low-rank and simple sparsity constraints. We speculate that they underperformed in this regime because the data is not exactly low-rank (as our temporal dimension is already small) as well as the sparsifying transforms (temporal FFT for L+S and temporal gradient for kt-SLR) lack adaptability to data compared to CNN and DLTG. The visualisation of reconstruction from 9-fold undersampling is shown in Fig. \ref{fig:vs_dltg1}, including the reconstruction from the CNN without data sharing and DLTG. The reconstructions of kt-SLR and L+S were omitted as their quantitative error were already much worse. One can see that, as with the 2D case, at aggressive undersampling rate dictionary-learning based method produced blocky artefacts, whereas the CNN methods were capable of reconstructing finer details (indicated in red ellipse). On the other hand, for the CNN without data sharing, one can notice grainy noise-like artefacts. Even though it was able to reconstruct the underlying anatomy more faithfully than DLTG, the overall error was worse. However, this artefact was not present in the images reconstructed by the CNN with data sharing. Although the quantitative result is not shown, CNN without data sharing in fact outperformed DLTG for low acceleration factor (3x) but not for more aggressive undersampling factor. This suggests that when the aliasing is severe, more drastic transformation is required, in which case for CNN to do better, we either need to increase depth, which would increase its computation cost, or increase the training samples. This confirms the importance of data sharing and the necessity to exploit the domain knowledge to simplify the learning problem for the case when the data is limited. Temporal profiles from the reconstructions are shown in Fig. \ref{fig:vs_dltg2}. Even though the data sharing itself results in data inconsistency in highly dynamic regions, the CNN was able to rectify this internally and reconstructed the correct motion with errors smaller than the other methods. This suggests the CNN's capability solve the joint de-aliasing and implicit estimation of dynamic motion.

\paragraph{Reconstruction with Noise}

\begin{figure}[t]
  \centering
  \includegraphics[width=0.45\textwidth]{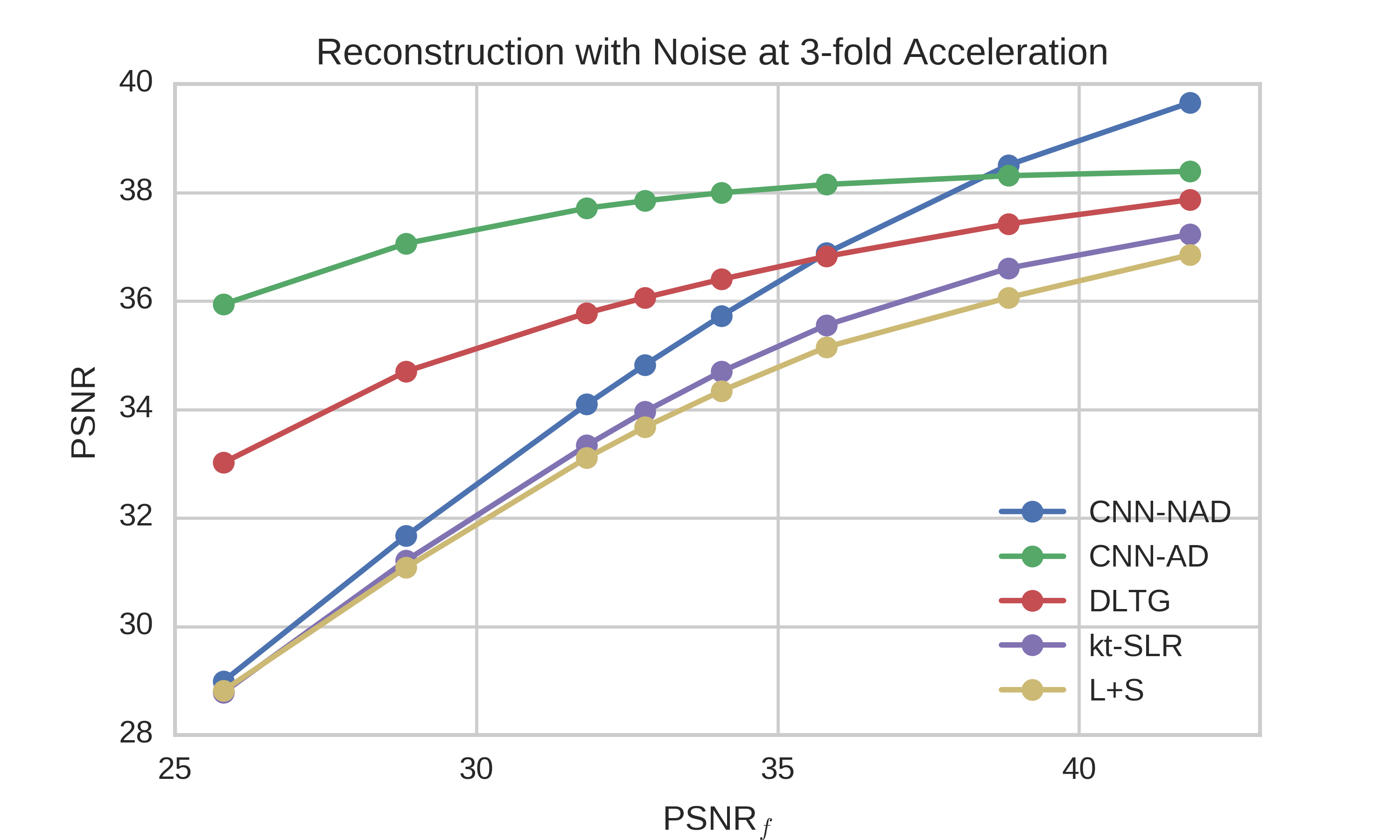}
  \caption{The aggregated test error across 10 subjects with injected noise. For different value of input noise power, PSNR$_f$ is shown. The corresponding reconstruction PSNR for CNN-NAD, CNN-AD, DLTG, kt-SLR and L+S are shown.}
\label{fig:noise}
\end{figure}

This section analyses the impact of acquisition noise in reconstruction performance. In this experiment we fixed the acceleration factor to be 3 and varied the level of noise in the data. Specifically, we tested for noise power $\sigma^2 \in [10^{-9}, 4\times 10^{-8}]$. For fully-sampled reconstruction, the noise power is equivalent to peak signal-to-noise (PSNR) values of $41.84$ dB and $25.81$ dB for $10^{-9}$ and $4\times 10^{-8}$ respectively, where PSNR was calculated as $10 \log_{10}(1/\text{MSE})$. The result is summarised at Fig \ref{fig:noise}, where we aggregate the reconstruction error from all 10 subjects. The input level of noise is indicated by PSNR$_f$ and for consistency, the reconstruction results are also indicated by PSNR (higher the better). For DLTG, we used the value $\lambda = 5\times 10^{-6}$ as recommended in \cite{Caballero2014}. DLTG showed decent robustness to noise, owing to the nature of underlying K-SVD, which has the effect of sparse coding denoising. For kt-SLR and L+S, we used the same parameters as before. They showed some robustness for small noise but they did not perform well in the presence of aggressive noise, as the implementations (and the data consistency step in particular) do not explicitly account for them. Changing such implementation is likely to improve the result.

\begin{figure*}[t]
  \centering
  \includegraphics[width=\textwidth]{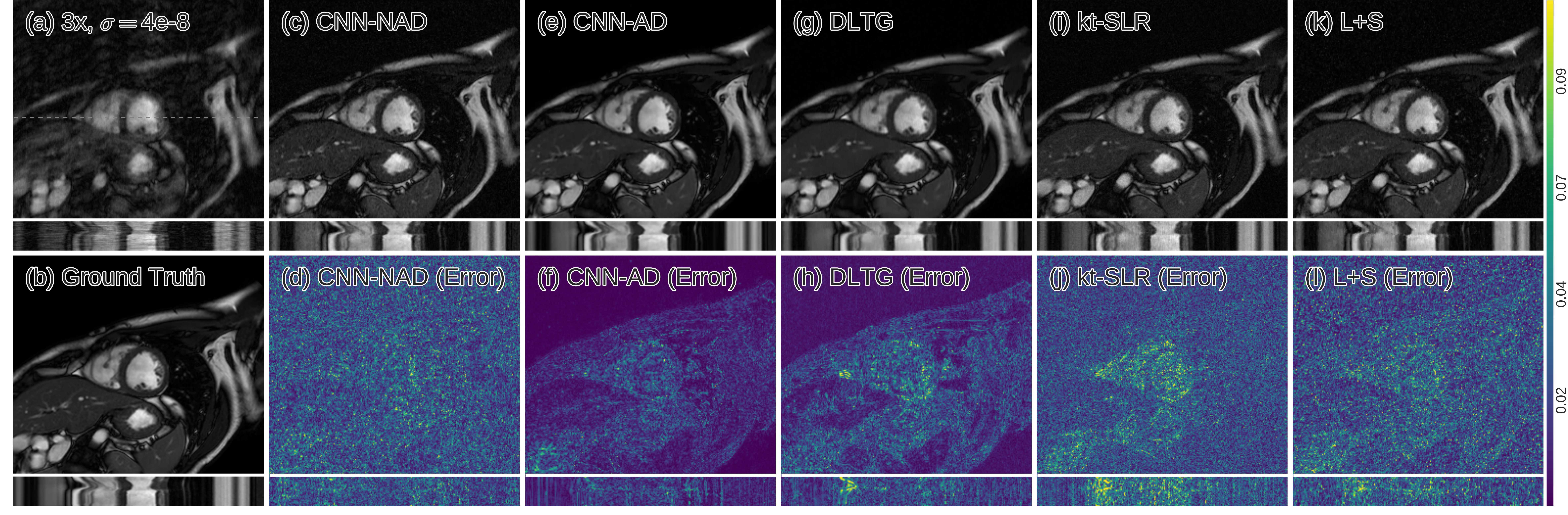}
  \caption{The reconstruction with noise $\sigma^2 = 4 \times 10^{-8}$. The aggregated test error across 10 subjects with injected noise. For different value of input noise power, PSNR$_f$ is shown. The corresponding reconstruction PSNR for CNN, finetuned CNN, DLTG, kt-SLR and L+S are shown.}
\label{fig:noise_rec}
\end{figure*}

For CNN, we used the model \emph{D5-C10(S)} as before and tested the following two variations. Firstly, we tested the performance of CNN from the previous section, which were trained in the absence of noise, denoted as \emph{CNN-NAD} (blue curve). It can be seen that for the low level of noise (PSNR $>35$ dB), CNN-NAD were able to maintain similar performance as the rest of the methods. However, the performance degraded almost at the same rate as kt-SLR and L+S for the high level of noise. We then trained CNN-NAD to adapt for noise as following. Firstly, we added noise in training data, where we randomly sample the noise power in the range $[10^{-9}, 4\times10^{-8}]$. Secondly, we modified our data consistency layers to account for noise. In particular, we initialised $\lambda$ for each DC layer as $\lambda = q / \sigma = 0.025$ (as in DLTG), made the parameters trainable. We trained the network for $3\times 10^4$ backpropagations and the result is denoted as \emph{CNN-AD} (green curve). Interestingly, the performance for very small noise ($>38$ dB) became worse compared to the original CNN. However, for further acceleration, it showed significant improvement for all level of noise, showing better robustness compared to other methods. We also observed that after fine-tuning, $\lambda$ was increased to $0.5$. This signifies that DLTG and CNN, even though the reconstruction framework shows similarity in terms of the iterative nature, are fundamentally different approaches and the required parameters also vary. Note that since we trained the network for a wide range of noise, the performance is likely to be improved if a narrower range of noise is selected for training. In practice, measuring the level of noise a-priori is non-trivial. However, our CNN showed the adaptability to the pre-specified range which indeed can be simulated in practice.

\paragraph{Reconstruction speed} Similar to the 2D case, the DLTG takes 6.6 hours per subject on CPU. For the CNN, each sequence was reconstructed on average $8.21s \pm 0.02s$ on GPU GeForce GTX 1080. This is significantly slower than reconstructing 2D images as introducing a temporal axis greatly increases the computational effort of the convolution operations. Nevertheless, the reconstruction speed of our method is much faster than DLTG and is reasonably fast for offline reconstruction. 		

\subsection{Memory Requirement} 

The memory requirement of the CNNs is based on the number of the network parameters, the number and the sizes of the intermediate activation maps and the space needed for computing the layer operations. The total number of the network parameters is simply given by the sum of all the layer parameters. Each convolution layer has $(k_x k_y k_t n_f' + 1) n_f$ parameters, where $k_x, k_y, k_t$ are the kernel sizes along $x$, $y$ and $t$, $n_f'$ and $n_f$ are the number of features of the incoming and current convolution layers respectively and one for the bias. For each DC layer, we also store one parameter for $\lambda$. For 2D reconstruction ($k_t=1$) and \emph{D5-C5} has about 0.6 million parameters, which occupies 2.3MB of the storage assuming single-precision floating point is used ($N_\text{precision}=4$ bytes). For dynamic reconstruction, \emph{D5-C10(S)} has 3.4 million parameters, which occupies about 13.6MB. 

At the training stage, more than three times of the number of parameters are required for computing the gradient. In addition, all the intermediate activation maps need to be stored to perform the backpropagation efficiently. For the proposed architecture, the most of the activation maps are of the convolution layer $C_i$'s; hence, the sum can be roughly estimated by $N_\text{batch} N_x N_y N_t N_f n_c (n_{d}-1) N_\text{precision}$.  With the input size $N_\text{batch} \times N_x \times N_y \times N_t = 1\times 256 \times 256 \times 1$, the memory required for the activation maps of \emph{D5-C5} is 335MB. For the dynamic models, the memory requirement further increases by the size of the temporal dimension $N_t=30$. Therefore, the aforementioned trick of cropping the images along $N_y$ is necessary to fit the model. For \emph{D5-C10(S)}, with the input size $1\times 256 \times (256/8) \times 30$, 2.4GB is required for storing the activation maps alone. Finally, to obtain the total memory consumption for the training stage, this value needs to be further multiplied by factors based on the implementation of backpropagation, operations including convolution and FFT as well as any compilation optimisation performed by the library. For example, most implementations of backpropagation require twice the value above accounting for forward- and backward- passes. We report that for our Theano implementation of \emph{D5-C10(S)}, the largest mini-batch size we could fit for the given input size on GeForce GTX 1080 (8GB) was 1. 

At the testing stage, the memory requirement is much less because the intermediate activation maps do not need to be stored if only the forward pass needs to be performed. In this case, the memory overhead is only the single largest activation map, which is $C_i$, scaled by implementation-specific factors. Note that as aforementioned, the patch extraction cannot be used at test time. Nevertheless, we did not observe any problem using \emph{D5-C10(S)} for input size $1\times 256\times 256 \times 30$ on GeForce GTX 1080. 

\section{Discussion and Conclusion}

In this work, we evaluated the applicability of CNNs for the challenge of reconstructing undersampled cardiac MR image
data. The experiments presented show that using a network with interleaved data consistency stages, it is feasible to obtain a model which can reconstruct images well. The CS and low-rank framework offers a mathematical guarantee for the
signal recovery, which makes the approach appealing in theory as well as in practice even though the required sparsity cannot generally be genuinely achieved in medical imaging. However, even though this is not the case for CNNs, we have empirically shown that a CNN-based approach can outperform them. In addition, at very aggressive undersampling rates, the CNN method was capable of reconstructing most of the anatomical structures more accurately based on the learnt priors, while classical methods do not guarantee such behaviour. 	

Note that remarkably, we were able to train the CNN on the small dataset. We used several strategies to alleviate the issue of overfitting: firstly, as we employed the iterative architecture, each subnetwork has relatively small receptive field. As a result, the network can only performs local transformations. Secondly, we applied intensive data augmentation so the network practically sees constantly sees a variation of the input, which makes it more difficult to overfit to any specific patterns. However, we speculate that given more training data, we can drop the data augmentation and let the network learn coarse features by incorporating, for example, dilated or strided convolution, which could further improve the performance.

It is important to note that in the experiments presented the data was produced by retrospective undersampling of back transformed complex images (equivalent to single-coil data) obtained through an original SENSE reconstruction. Although the application of CNN reconstruction needs to be investigated in the more practical scenario of full array coil data from parallel MR, the results presented show a great potential to apply deep learning for MR reconstruction. The additional richness of array coil data has the potential to further improve performance, although it will also add considerable complexity to the required CNN architecture. 

In this work, we were able to show that the network can be trained using arbitrary Cartesian undersampling masks of fixed sampling rate rather than selecting a fixed number of undersampling masks for training and testing. In addition, we were able to pre-train the network on various undersampling rates before fine-tuning the network. This suggests that the network was capable of learning a generic strategy to de-alias the images. A further investigation should consider how tolerant the network is for different undersampling patterns such as radial and spiral trajectories. As these trajectories provide different properties of aliasing artefacts, a further validation is appropriate to determine the flexibility of our approach. However, radial sampling naturally fits well with the data sharing framework and therefore can be expected to push the performance of the network further. The data sharing approach may also make it feasible to adopt regular undersampling patterns which are intrinsically more efficient.  Another interesting direction would be to jointly optimise the undersampling mask using the learning framework. 

To conclude, although CNNs can only learn local representations which should not affect global structure, it remains to be determined how the CNN approach operates when there is a pathology present in images, or other more variable content. We have performed a cross-validation study to ensure that the network can handle unseen data acquired through the same acquisition protocol. Generalisation properties must be evaluated carefully on a larger dataset. However, CNNs are flexible in a way such that one can incorporate application specific priors to their objective functions to allocate more importance to preserving features of interest in the reconstruction, provided that such expert knowledge is available at training time. For example, analysis of cardiac images in clinical settings often employs segmentation and/or registration. Multi-task learning is a promising approach to further improve the utility of CNN-based MR reconstructions.

\section*{Acknowledgment}

The work was partially funded by EPSRC Programme Grant (EP/P001009/1).

\ifCLASSOPTIONcaptionsoff
  \newpage
\fi

\IEEEtriggeratref{33}
% \IEEEtriggercmd{\enlargethispage{-5in}}


\bibliographystyle{IEEEtran}
\end{document}